\documentclass{article}

\usepackage[preprint]{neurips_2025}

\usepackage[utf8]{inputenc} 
\usepackage[T1]{fontenc}    
\usepackage{hyperref}       
\usepackage{url}            
\usepackage{booktabs}       
\usepackage{amsfonts}       
\usepackage{nicefrac}       
\usepackage{microtype}      
\usepackage{xcolor}         
\usepackage{tabularx, multirow, multicol, graphicx, makecell, enumitem, color, soul, colortbl, adjustbox, amsmath, subcaption, pifont, amssymb}
\usepackage{authblk}

\usepackage{floatrow}
\newfloatcommand{capbtabbox}{table}[][\FBwidth]

\usepackage{algorithm}
\usepackage{algpseudocode}

\title{UniCode²: Cascaded Large-scale Codebooks for Unified Multimodal Understanding and Generation}

\author{Yen-chieh Chan\thanks{All authors marked with \textsuperscript{$\ast$} are co-first authors.}\;\,, Huasong Zhong\textsuperscript{$\ast$}, Yan Li\textsuperscript{$\ast$}\thanks{Project Leader}\;, Zhenheng Yang\thanks{Corresponding Author}}
\affil{ByteDance}

\begin{document}
\bibliographystyle{plain}

\maketitle

\begin{abstract}


Unified multimodal large language models (MLLMs) have shown promise in jointly advancing multimodal understanding and generation, with visual codebooks discretizing images into tokens for autoregressive modeling. Existing codebook-based methods either rely on small vocabularies ($\approx$16K entries) that lack fine-grained semantics or naïvely scale up, resulting in low token utilization and unstable training.
We propose \textbf{UniCode²}, a cascaded codebook framework enabling large-scale, semantically aligned, and stable visual tokenization. By clustering millions of SigLIP sequence embeddings, we build a 500K-entry codebook that preserves vision-language alignment while expanding capacity. Stability is ensured via a cascaded design: a frozen codebook anchors the embedding space, and a trainable codebook refines task-specific semantics. This decoupling promotes high utilization and robust learning. 
Moreover, the alignment of our visual tokens with textual semantics enables seamless integration with pretrained diffusion decoders, supporting high-quality visual synthesis with minimal adaptation.
UniCode² delivers strong performance across diverse benchmarks, demonstrating the viability of scaling visual token spaces without sacrificing stability, semantics, or modularity.

\end{abstract}

\section{Introduction}

Multimodal large language models (MLLMs) have emerged as a central paradigm for bridging vision and language, enabling tasks such as visual question answering, text-to-image generation, and multimodal reasoning~\cite{wang2024qwen2, achiam2023gpt, chen2024expanding, team2023gemini}. Within this space, Unified MLLMs have gained traction for jointly addressing understanding and generation within a single framework~\cite{xie2024show, zhang2025unified, dong2023dreamllm, ma2024janusflow, tong2024metamorph}. Notably, codebook-based approaches leverage discrete visual tokens to facilitate a unified autoregressive modeling paradigm for vision-language tasks~\cite{liu2024world, ma2025unitok, qu2024tokenflow, wu2024liquid}.

\begin{figure}[htbp]
  \centering
  \begin{subfigure}{0.50\textwidth}
    \includegraphics[width=\textwidth]{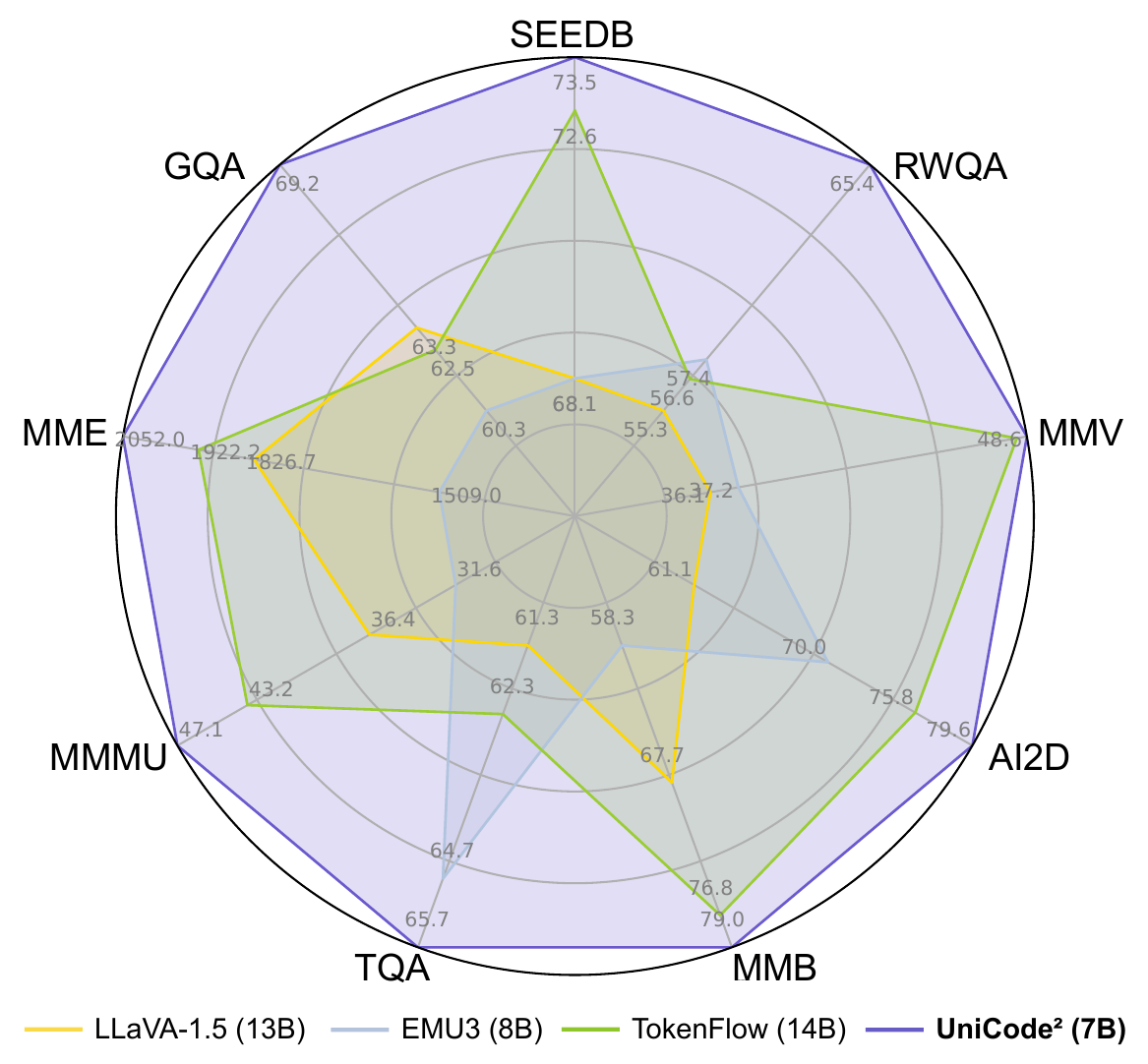}
    \caption{Understanding benchmarks comparison.}
    \label{fig:subfig1}
  \end{subfigure}
  \hfill
  \begin{subfigure}{0.48\textwidth}
    \includegraphics[width=\textwidth]{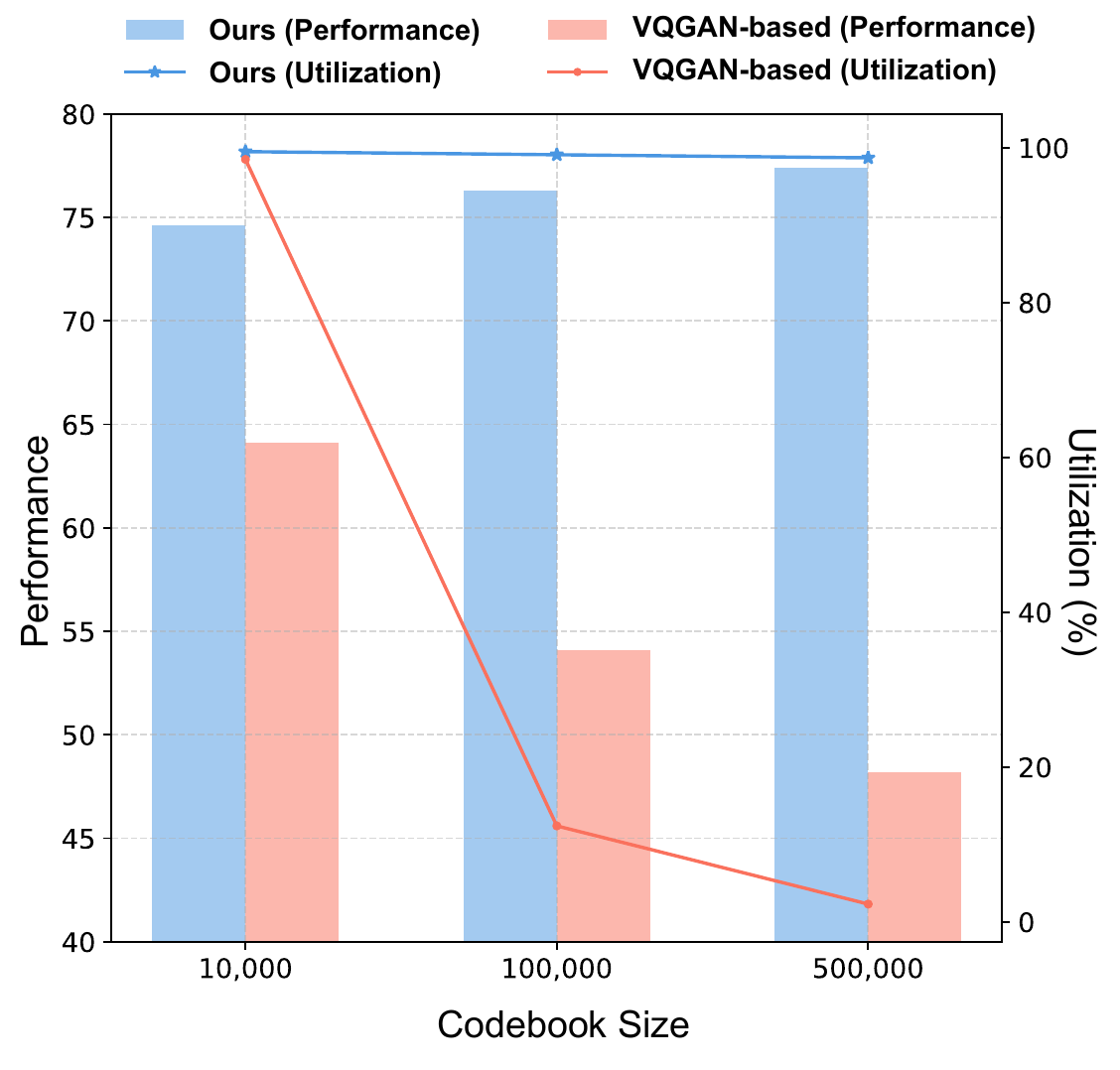}
    \caption{Codebook size vs. performance \& utilization.}
    \label{fig:subfig2}
  \end{subfigure}
  \vspace{-2mm}
  \caption{(a) \textbf{Comparison} among four models across various understanding benchmarks.  
    (b) \textbf{Motivation}: existing methods suffer sharp drops in performance and utilization as codebook size grows; our proposed UniCode² maintains \textasciitilde99\% utilization and strong performance.}
  \label{fig:mainfig}
\end{figure}

Codebook-based methods~\cite{van2017neural, wang2024image, esser2021taming, lee2022autoregressive} have shown pixel-level generation capabilities but remain limited in visual understanding~\cite{bao2021beit, wu2024vila, wang2024image}. Existing solutions mainly pursue two directions.
The first focuses on \textbf{scaling codebook size }to improve expressiveness and granularity~\cite{zhu2024scaling}. However, partial update schemes—updating only a subset of active codes per step~\cite{shi2024taming, yu2023language}—cause a widening gap between unused codes and the encoder’s output, leading to under-utilization and unstable training~\cite{zhu2024scaling}. As a result, even advanced approaches~\cite{shi2024taming, zheng2023online} remain limited to under 100K entries with restricted utilization.
The second line of research aims to \textbf{improve codebook semantic quality}~\cite{ma2025unitok}. Most visual tokenizers, being randomly initialized and optimized for reconstruction, capture low-level appearance cues but fail to encode high-level semantics crucial for cross-modal understanding~\cite{xie2024show, zhou2024transfusion}. To address this, some methods~\cite{wu2024vila, wang2024image} incorporate contrastive learning to align token semantics with language. However, unified tokenizers optimized with multimodal objectives often struggle to converge compared to domain-specific models~\cite{ma2025unitok, wu2024vila}. Others~\cite{wu2024janus, qu2024tokenflow, chen2025semhitok, huang2025illume} attempt to model semantics and pixels using dual-codebook designs, but this adds complexity to training and inference.

These limitations underscore the core challenges: existing visual tokenizers are either semantically weak or structurally brittle when scaled. This raises a fundamental question—\emph{\textbf{is it possible to build a truly large-scale codebook that is semantically rich, stably trainable, and highly utilized}}—thus enabling strong performance for both understanding and generation within a unified framework?

The limitations above call for rethinking the objective of codebook construction—from mere compression to a mechanism for semantic abstraction and cross-modal alignment. We contend that a \textbf{high-capacity, semantically grounded codebook is both necessary and achievable}. Leveraging SigLIP~\cite{zhai2023sigmoid}'s sequence-level embeddings, which encode contextualized visual semantics aligned with language, we cluster millions of unlabeled images to build a \textbf{500K-entry codebook}. This approach combines rich semantic priors with a decoupled construction, enabling stable, high-coverage tokenization without collapse. Thus, the codebook transforms from a bottleneck into a bridge linking vision and language, pretraining and generation, and ultimately understanding and synthesis.

We introduce \textbf{UniCode²}, a unified framework leveraging a Cascaded Large-scale Codebook for multimodal understanding and generation. UniCode² employs a two-stage codebook architecture sharing a 500K-entry vocabulary derived from clustering millions of SigLIP sequence embeddings. The first-stage codebook is frozen, providing a stable, semantically aligned embedding space, while the second-stage codebook is trainable, enabling task-specific refinement. This decoupling of indexing and adaptation ensures high code utilization and stable optimization, addressing common collapse and underutilization issues in large-scale codebook training.
Unlike conventional tokenizers focused on pixel reconstruction, UniCode²'s token space is explicitly grounded in semantic SigLIP features, capturing abstract concepts essential for vision-language tasks and yielding superior understanding performance. Its modular design also supports seamless integration with pretrained diffusion decoders (e.g., SDXL~\cite{podell2023sdxl}, FLUX~\cite{blackforestlabs2024}). For example, combined with FLUX, UniCode² attains state-of-the-art results by training only a lightweight mapping on 200K images. This clear separation of understanding and generation reduces training costs and enhances transferability across modalities.

Our contributions are summarized as follows:
\begin{itemize}[leftmargin=*]
    \item \textbf{Large-scale SigLIP-Semantic Codebook}: We cluster millions of SigLIP embeddings to build a 500K-entry visual codebook that achieves scalability, high utilization, and vision-text alignment, effectively mitigating the limitations of existing codebook methods in understanding tasks.
    \item \textbf{Stable Training via Cascaded Codebooks}: We propose a cascaded codebook architecture that separates indexing and refinement, enabling stable training and rich semantic representation for fine-grained visual concepts even under large-scale vocabularies.
    \item \textbf{Plug-and-Play Synthesis}: Our semantically grounded token space seamlessly integrates with pretrained diffusion decoders, enabling high-fidelity synthesis with minimal additional training.
    \item \textbf{Extensive Emperimental Validation}: Through comprehensive experiments across multimodal benchmarks, we demonstrate that UniCode² consistently outperforms existing methods.
\end{itemize}

\section{Related Work}

\subsection{Unified Multimodal Large Language Models}
Recent work seeks to unify vision understanding and generation within a single MLLM~\cite{zhang2025unified, xie2024show,song2025dualtoken, wang2024emu3, xiao2024omnigen}. Show-O~\cite{xie2024show} and Chameleon~\cite{team2024chameleon} achieve token-level fusion for joint modeling with minimal architectural changes, while MetaMorph~\cite{tong2024metamorph} enables generative capabilities via instruction tuning alone.  
Codebook-based MLLMs offer an alternative by discretizing images into tokens compatible with LLMs~\cite{dong2023dreamllm, sun2024generative, ge2024seed, lin2025toklipmarryvisualtokens}. JanusFlow~\cite{ma2024janusflow} decouples modality-specific encoders with rectified flow for joint modeling, while Janus~\cite{wu2024janus} and TokenFlow~\cite{qu2024tokenflow} refine similar paradigms. UniTok~\cite{ma2025unitok} introduces multi-codebook quantization to enhance token expressiveness, and VILA-U~\cite{wu2024vila} adopts a unified decoder but inherits limitations from fixed tokenizers.  
UniCode² aligns token space with SigLIP-level semantics, capturing abstract visual concepts for strong understanding performance. Its modular design further enables plug-and-play generation with pretrained diffusion decoders.

\subsection{Codebook Design for Visual Understanding}
Recent advances in codebook-based visual tokenization have improved tasks like image classification and visual question answering~\cite{awais2025foundation, gui2024survey, xiong2024efficientsam, chenmai, fang2023eva, zhao2025qlip}. BEiT~\cite{bao2021beit} introduced masked image modeling by quantizing visual inputs into discrete tokens, enabling transformer architectures to process images like language~\cite{wang2023image, peng2022beit, fang2024eva, wang2023image}. 
Concurrently, VQToken~\cite{zhang2025token} proposes a learned codebook for discretizing vision embeddings, focusing on efficient high-compression token reduction for video-language models.
Selftok~\cite{wang2025discrete} integrated autoregressive priors and diffusion processes, enhancing comprehension and generation without paired text-image training. These developments demonstrate the versatility of codebooks in unifying discrete representations across vision and language.

\subsection{Codebook Design for Visual Generation}
Discrete representations also underpin modern visual generation~\cite{tian2024visual,zhang2025unified, ren2025beyond, sun2023emu, zhu2024addressing, chern2024anole}. Early works such as VQVAE~\cite{van2017neural} and VQGAN~\cite{esser2021taming} encode images into compact token sequences, enabling autoregressive generation or diffusion-based decoding~\cite{cao2023efficient, ma2025unitok, wang2024omnitokenizer}. Extensions like ViT-VQGAN~\cite{yu2021vector} improve fidelity, but reconstruction-oriented training often leads to semantically misaligned tokens, hindering cross-modal controllability.
To overcome these limitations, recent approaches scale up codebooks and enforce semantic regularization. FQGAN~\cite{bai2024factorized} introduces disentangled multi-codebook designs to enhance diversity and reduce redundancy. VQGAN-LC~\cite{zhu2024scaling} scales the codebook to 100K entries, leveraging CLIP-based initialization and projection for better semantic alignment. 

\section{Approach}

\begin{figure*}[t!]
  \centering
  \includegraphics[width=1.0\linewidth]{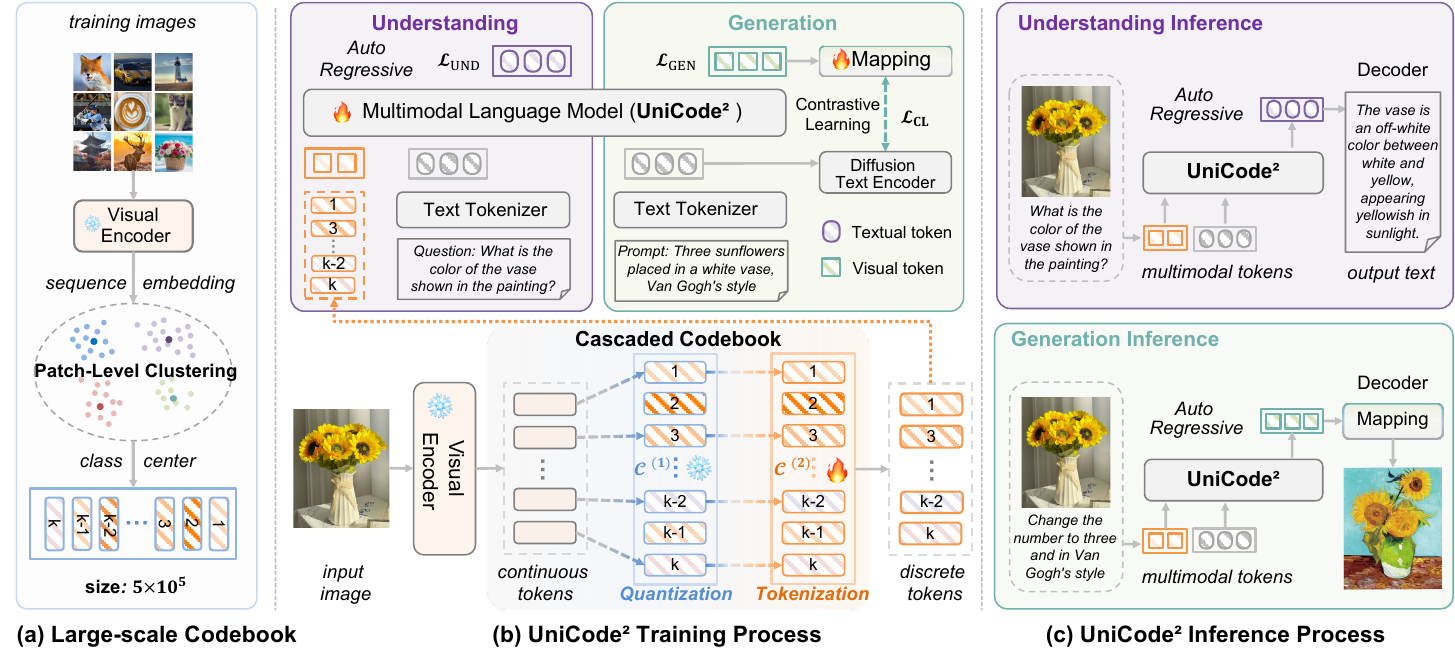}
  \caption{UniCode² framework overview. (a) High-capacity codebook initialized by clustering large-scale patch embeddings. (b) During training, visual inputs are tokenized via cascaded codebooks and optimized via unified autoregressive and contrastive objectives. (c) Inference supports both understanding and generation, with visual tokens seamlessly decoded via diffusion models.}
  \label{approach}
\end{figure*}

We begin by motivating the design of a large-scale codebook (§\ref{sec_3_1}), then describe a cascaded quantization pipeline (§\ref{sec_3_2}) for obtaining stable and compact visual tokens. Next, we present a unified autoregressive framework for visual understanding (§\ref{sec_3_3}) and generation (§\ref{sec_3_4}), followed by the overall training and inference strategy (§\ref{sec_3_5}).

\subsection{Motivation: Toward Optimal Codebook Design}
\label{sec_3_1}

A well-constructed visual codebook is fundamental to bridging continuous vision signals and discrete language modeling. To ensure that the learned codebook captures diverse and meaningful visual semantics while maintaining high usage efficiency, we propose a principled formulation that quantifies codebook quality from two perspectives: \textbf{semantic alignment} and \textbf{utilization regularization}.

Given visual embeddings $\mathcal{E} = \{\mathbf{e}_1, \dots, \mathbf{e}_N\}$, each $\mathbf{e}_i$ is assigned to its nearest codeword $\mathbf{c}_k$ in a codebook $\mathcal{C} = \{\mathbf{c}_1, \dots, \mathbf{c}_K\}$.
An effective codebook should ensure that each embedding is well-aligned to its assigned codeword, while also encouraging uniform utilization of codewords across the dataset. We formalize this intuition via the following objective:
\begin{equation}
\mathcal{L}_{\text{codebook}} = 
\underbrace{\sum_{k=1}^{K} \sum_{\mathbf{e}_i \in \mathcal{C}_k} \left\| \mathbf{e}_i - \mathbf{c}_k \right\|^2}_{\text{semantic alignment}} 
+ \lambda \underbrace{\sum_{k=1}^{K} \left\| \mathbf{c}_k - \bar{\mathbf{c}} \right\|^2}_{\text{utilization regularization}},
\label{eq_1}
\end{equation}
where $\bar{\mathbf{c}}$ is the codebook centroid, and $\lambda > 0$ controls the balance between fidelity and spread.

The first term enforces \emph{semantic alignment}, encouraging each embedding to be close to its assigned centroid, thereby preserving fine-grained semantics in the quantization process. The second term promotes \emph{utilization regularization}, penalizing collapse by encouraging all codewords to spread around the embedding space evenly. This dual objective ensures that the learned codebook avoids degenerate solutions—such as assigning all inputs to a small subset of codewords—and instead fosters high-entropy, high-coverage quantization.

To further illustrate the necessity of both objectives, we note that the expected reconstruction error can be lower-bounded by the average intra-cluster variance:
\begin{equation}
\mathbb{E}_{\mathbf{e}_i \sim p(\mathbf{e})} \left[ \| \mathbf{e}_i - Q(\mathbf{e}_i) \|^2 \right] 
\geq \frac{1}{K} \sum_{k=1}^{K} \text{Var}[\mathbf{e}_i \mid \mathbf{e}_i \in \mathcal{C}_k],
\end{equation}
where $Q(\mathbf{e}_i)$ denotes the quantized codeword for $\mathbf{e}_i$. This bound is tightest when each cluster is compact (low intra-cluster variance) and equally populated, directly reflecting the goals of semantic alignment and balanced utilization.
Moreover, from an information-theoretic perspective, maximizing the entropy of codeword assignments:
\begin{equation}
H(q) = -\sum_{k=1}^{K} q(k) \log q(k),
\end{equation}
where $q(k)$ is the empirical probability of selecting codeword $\mathbf{c}_k$, yields maximal representational diversity. Our formulation implicitly encourages high entropy through the utilization term, thereby balancing \textit{semantic compactness} and \textit{codeword diversity}, which are both crucial for downstream reasoning and generation tasks.

\subsection{Semantic Codebook Construction via Clustering}
\label{sec_3_2}

The objective in Eq.~\ref{eq_1} naturally aligns with the classical clustering formulation, where minimizing the sum of squared distances between embeddings and cluster centers directly corresponds to the semantic alignment term. Meanwhile, the utilization regularization term encourages balanced cluster sizes, preventing collapse into a few dominant clusters. By performing large-scale clustering on pretrained SigLIP embeddings—which already encode rich, semantically meaningful visual features aligned with language supervision—we effectively initialize a codebook that satisfies both semantic fidelity and utilization balance.

Given a large corpus of natural images $\mathcal{D}=\{I_i\}_{i=1}^N$, we extract dense patch-level embeddings using a pretrained SigLIP model~\cite{zhai2023sigmoid}. For each image $I_i$, the encoder $\phi:I \rightarrow \{x_j^{(j)}\in \mathbb{R}^d \}_{j=1}^{L}$ maps the image into $L$ patch embeddings of dimension $d$. Aggregating all patches from all images yields a corpus of embeddings $\mathcal{X} = \cup_{i=1}^{N}\{x_i^{(j)}\}_{j=1}^L$.

We perform $K$-means clustering over $\mathcal{X}$ to obtain $K=500,000$ centroids $\mathcal{C} = \{\mathbf{c}_k\}_{k=1}^K \subset \mathbb{R}^d$. This massive vocabulary ensures high coverage of diverse visual semantics. Importantly, since the clustering is performed over SigLIP embeddings, each centroid implicitly corresponds to a semantically grounded visual concept. The resulting codebook $\mathcal{C}$ serves as a semantic foundation for subsequent image quantization and tokenization.

In the following section, we describe how this codebook is leveraged within a cascaded architecture to further enhance flexibility and expressiveness.

\subsection{Cascaded Codebook Architecture}
\label{sec_3_3}


Building upon the semantic codebook $\mathcal{C}$ obtained in Section~\ref{sec_3_2}, we introduce a \emph{cascaded codebook architecture} to address the challenges of stability and adaptability in large-scale visual tokenization. This architecture decouples the stable discretization provided by a \emph{frozen quantization codebook} from downstream task-specific semantic adaptation, achieved via a \emph{learnable tokenization codebook}.

Given an input image $x$, we extract its patch-level sequence representation using a frozen SigLIP image encoder $f_{\rm SigLIP}$, yielding a sequence of continuous visual tokens:
\begin{equation}
    \mathbf{z} = f_{\rm SigLIP}(I) = [\mathbf{z}_1, \mathbf{z}_2, ..., \mathbf{z}_T], \mathbf{z}_i\in\mathbb{R}^d,
\end{equation}
where $T$ is the number of image patches and $d$ is the feature dimension. This continuous sequence $\mathbf{z}$ undergoes a two-stage vector quantization process: the first stage produces stable quantized discrete tokens, while the second stage generates learnable task-specific token embeddings.

In the first stage, we employ the \emph{frozen quantization codebook} $\mathcal{C}^{(1)}$, which is directly inherited from $\mathcal{C}$ constructed in Section~\ref{sec_3_2}. This codebook remains fixed throughout training and is used to quantize each patch embedding $\mathbf{z}_i$ into a discrete token:
\begin{equation}
    v_i = \arg\min_k \|\mathbf{z}_i - \mathbf{c}_k^{(1)}\|_2,
\end{equation}
where $\mathbf{c}_k^{(1)}$ denotes the $k$-th codeword in the frozen codebook $\mathcal{C}^{(1)}$. This step discretizes continuous features into a stable token space $\mathbf{v} = [v_1, ..., v_T]$, preserving the geometric and semantic structure of the pretrained embedding manifold.

To enable task-specific adaptation and enrich semantic expressiveness, we introduce a \emph{learnable tokenization codebook} $\mathcal{C}^{(2)}$ in the second stage. This codebook is initialized from the same codebook $\mathcal{C}$ constructed in Section~\ref{sec_3_2}, but unlike $\mathcal{C}^{(1)}$, it is updated dynamically during training. $\mathcal{C}^{(2)}$ functions as a learnable visual embedding layer, mapping discrete visual tokens into a continuous visual embedding space:
\begin{equation}
    E_{\rm vis}(v_i) = \mathbf{c}_{v_i}^{(2)},
\end{equation}
transforming the input image $I$ into the visual embedding sequence $[E_{\rm vis}(v_1), ..., E_{\rm vis}(v_T)]$.

\subsection{Multimodal Autoregressive Modeling}
\label{sec_3_4}

Having obtained the visual embedding sequence $[E_{\rm vis}(v_1), ..., E_{\rm vis}(v_T)]$, we aim to integrate the visual modality into the LLM framework. In LLMs, textual tokens $\mathbf{q} = [q_1, \dots, q_M]$ are typically processed through a text embedding layer $E_{\rm text}$, resulting in a textual embedding sequence $\mathbf{x}_{\rm text} = [E_{\rm text}(q_1), ..., E_{\rm text}(q_M)]$. However, the feature dimensions of the visual and textual embeddings are not directly compatible. To address this, we introduce an MLP projector $f_{\rm proj}$ to align the visual embeddings with the textual embeddings: $\mathbf{x}_{\rm vis} = f_{\rm proj}\big([E_{\rm vis}(v_1), ..., E_{\rm vis}(v_T)]\big)$. The concatenated multimodal input $\mathbf{x} = [\mathbf{x}_{\rm vis}, \mathbf{x}_{\rm text}]$ is then fed into the LLM backbone, which autoregressively models the conditional distribution over the next text token for understanding tasks such as VQA.

\begin{equation}
    \mathcal{L}_{\rm UND} = - \sum_{t=1}^M \log P(q_t \mid \mathbf{x}_{\rm vis}, q_{<t}; \theta).
\end{equation}

\subsection{Unified Generation via Diffusion-ready Visual Tokens}
\label{sec_3_5}

While the previous section focuses on understanding tasks that map from image to text, our architecture naturally supports the inverse process of text-to-image generation by leveraging the same visual token space. The core idea is to autoregressively predict discrete visual tokens $\mathbf{v} = [v_1, ..., v_T]$ from an input prompt $\mathbf{p} = [p_1, ..., p_L]$, and subsequently decode these tokens into images via pretrained diffusion models. Given a prompt $\mathbf{p}$, we tokenize it via $E_{\rm text}$ and feed it into the LLM. The model is trained to generate the discrete visual tokens $\mathbf{v}$ one by one:
\begin{equation}
    \mathcal{L}_{\rm GEN} = -\sum_{t=1}^T{\log{P}(v_t| v_{<t}, \mathbf{p};\theta)}.
\end{equation}

To ensure that the generated tokens are compatible with downstream diffusion models, we introduce a learnable mapping function $\mathcal{M}_{\phi}:\mathbb{R}^d \rightarrow \mathbb{R}^{d^{\prime}}$ that projects each token embedding into the latent space expected by the diffusion decoder. Formally, for each generated token $v_t$, we compute:
\begin{equation}
    \widetilde{\mathbf{z}_t} = \mathcal{M}_{\phi}\big(E_{\rm vis}(v_t)\big),
\end{equation}
where $\widetilde{\mathbf{z}_t}$ serves as the conditioning input to the diffusion model. 

To align this mapping with the prompt semantics, we adopt a Contrastive Loss (CL) that encourages the mapped visual token sequence $\widetilde{\mathbf{z}} = [\widetilde{\mathbf{z}}_1, ..., \widetilde{\mathbf{z}}_T]$ to be semantically close to the prompt embedding $\mathbf{z}_\mathbf{p} = f_{\rm enc}(\mathbf{p})$, where $f_{\rm enc}$ is a frozen text encoder:
\begin{equation}
    \mathcal{L}_{\rm CL} = -\log{\frac{\exp{\big({\rm sim}{(\widetilde{\mathbf{z}}, \mathbf{z}_\mathbf{p})/ {\tau}}\big)}}{\sum_{j=1}^B{\exp{\big({\rm sim}{(\widetilde{\mathbf{z}}, \mathbf{z}_\mathbf{p}^{(j)})/ {\tau}}\big)}}}},
\end{equation}
where ${\rm sim}(\cdot,\cdot)$  denotes cosine similarity and $\tau$ is a temperature hyperparameter.

Finally, the mapped sequence $\widetilde{\mathbf{z}}$ conditions pretrained diffusion decoders for image synthesis, offering a plug-and-play interface between LLMs and generative models.

\subsection{Training and Inference}
\label{sec_3_6}

Our framework unifies vision-language understanding and generation within the autoregressive paradigm. Given paired image-text data $(\mathbf{I}, \mathbf{y})$, the image $\mathbf{I}$ is encoded by SigLIP and quantized into discrete tokens $\mathbf{v}=[v_1, ..., v_T]$ via the cascaded codebook.  
For \textbf{understanding} tasks, the model takes visual tokens $\mathbf{v}$ and a textual prompt $\mathbf{q}$ as input, autoregressively predicting the target sequence $\mathbf{y}$ with a language modeling loss $\mathcal{L}_{\rm UND}$.  
For visual \textbf{generation} tasks, the model generates visual tokens $\mathbf{\hat{v}}=[\hat{v}_1, ..., \hat{v}_T]$ from a textual prompt $\mathbf{p}$, supervised by a generation loss $\mathcal{L}_{\rm GEN}$. Predicted tokens are embedded and mapped via a module $\mathcal{M}_{\phi}$, trained with a contrastive loss $\mathcal{L}_{\rm CL}$ to align with encoder-derived prompt embeddings.  

At \textbf{inference}, the framework supports both directions seamlessly. For understanding, visual inputs are quantized into tokens $\mathbf{v}$, and the model generates text conditioned on $\mathbf{v}$ and queries. For generation, the model synthesizes visual tokens $\mathbf{\hat{v}}$ from a prompt $\mathbf{p}$, which are embedded, mapped via $\mathcal{M}_{\phi}$, and decoded by pretrained diffusion models. This modular design ensures compatibility by aligning the token space with the latent space of the decoders, enabling high-quality synthesis without fine-tuning.

\section{Experiment}

\subsection{Experimental Setup} 

\paragraph{Datasets.} To equip UniCode² with unified vision-language understanding and generation, we adopt a multi-stage curriculum learning paradigm inspired by recent MLLM practices~\cite{lu2024ovis, li2024llava, wang2024qwen2, bai2025qwen2}, leveraging the open-source corpus from LLaVA-OneVision~\cite{li2024llava}. 
To train the visual generation pathway, we augment 200K image-caption pairs from JourneyDB~\cite{sun2023journeydb}. These pairs supervise the contrastive alignment (§\ref{sec_3_4}), ensuring semantic consistency of visual tokens.

\paragraph{Implementation Details.} (1) \textit{Model Components:} We use Qwen2.5-7B-Instruct~\cite{yang2024qwen2} as the language backbone and siglip-so400M-patch14-384~\cite{zhai2023sigmoid} as the visual encoder for multimodal input and codebook construction. For image generation, we integrate FLUX.1-dev~\cite{blackforestlabs2024} and SDXL-base-1.0~\cite{podell2023sdxl}, using default settings (e.g., 50 denoising steps, guidance scale 3.5 for FLUX).  
(2) \textit{Codebook Construction:} The codebook is built by applying K-means clustering to sequence embeddings from 558K images used in Stage 1 of LLaVa-OneVision~\cite{li2024llava}, with 500K clusters as the target vocabulary size. 
(3) \textit{Training Setup:} Training is performed on 128 NVIDIA H100 GPUs with a per-GPU batch size of 4. AdamW~\cite{loshchilov2017decoupled} is used with learning rates of $1 \times 10^{-3}$ for Stage 1 and $1 \times 10^{-5}$ for later stages, applying a 0.03 warmup ratio.

\paragraph{Evaluation Metrics.} For multimodal understanding, we evaluate UniCode² on a comprehensive suite of vision-language benchmarks, including SEED-Bench~\cite{li2023seed}, MM-Vet~\cite{yu2023mm}, MMStar~\cite{chen2024we}, GQA~\cite{hudson2019gqa}, TextVQA~\cite{singh2019towards}, AI2D~\cite{kembhavi2016diagram}, RealWorldQA~\cite{realworldqa}, MMMU~\cite{yue2024mmmu}, MMBench~\cite{liu2024mmbench}, and MME~\cite{liang2024survey}.
For visual generation, we report results on GenEval~\cite{ghosh2023geneval} and DPO-Bench~\cite{hu2024ella}, focused on the semantic alignment and quality of text-conditioned image synthesis.
Following prior work such as TokenFlow~\cite{qu2024tokenflow}, we do not adopt FID scores, as they have been shown to exhibit weak correlation with human-perceived semantic fidelity and relevance in visual generation settings~\cite{chen2023pixart, podell2023sdxl, sun2023emu}.

\subsection{Quantitative Results}

\paragraph{Results on Understanding Tasks.}
We evaluate UniCode² on diverse understanding benchmarks, with results summarized in Table~\ref{table_und}. Key observations include: (\romannumeral1) UniCode² achieves state-of-the-art performance among discrete-input models, outperforming baselines like TokenFlow~\cite{qu2024tokenflow}, despite its larger 14B LLM. (\romannumeral2) UniCode², trained solely on open-source data, performs competitively with continuous-token MLLMs like Ovis~\cite{lu2024ovis}. (\romannumeral3) UniCode² excels across diverse tasks, from VQA to scientific and diagram reasoning, demonstrating robust understanding enabled by its high-capacity and semantically aligned codebook, narrowing the gap with continuous-input models.

\setlength{\tabcolsep}{3.5pt}
\begin{table*}[t]
\centering
\small
\adjustbox{max width=\textwidth}{
\begin{tabular}{l | c | c c c  c c c c c c c }
\toprule
\textbf{Method} & \textbf{\# Params} & \textbf{SEEDB} & \textbf{MMV} & \textbf{MMStar} & \textbf{GQA} & \textbf{TQA} & \textbf{AI2D} & \textbf{RWQA} & \textbf{MMMU} & \textbf{MMB} & \textbf{MME}   \\
\midrule
\multicolumn{12}{l}{\textcolor[RGB]{105, 105, 105}{\textit{Continuous Visual Input}}} \\
\midrule
InstructBLIP~\cite{liu2023visual} & 13B& 58.8 & 25.6 & 32.7 & 49.5 & 50.7 & -- & -- & 36.0 & -- & --   \\
LLaVA-1.5~\cite{liu2024improved}  &  13B & 68.1 & 36.1 & 33.1 & 63.3 & 61.3 & 61.1 & 55.3 & 36.4 & 67.7 & 1826.7   \\
ShareGPT4V~\cite{chen2024sharegpt4v} &  7B   & 69.7 &  37.6 & 35.7   & 63.3 & 60.4 & 58.0 & 54.9 & 37.2 & 68.8 & 1943.8  \\
NExT-GPT~\cite{wu2024next} &  7B   & 57.5 & -- & --  & -- & -- & -- & -- & -- & 58.0 & --    \\
Qwen-VL~\cite{bai2023qwen} &  7B  & 57.7 & -- & 37.5 & 57.5 & -- & -- & -- & -- & -- & 1848.3  \\
LLaVA-OV~\cite{li2024llava} & 7B & 75.4 & 57.5 & 61.7 & -- & -- & 81.4 & 66.3 & 48.8 & 80.8 & 1998.0 \\
Ovis~\cite{lu2024ovis} & 7B & -- & -- & 49.5 & -- & -- & -- & 57.9 & 44.7 & 77.4 & 1882.0 \\
Janus~\cite{wu2024janus} &  1.3B  & 63.7 & 34.3 & 37.6  & 59.1 & -- & -- & -- &30.5 & 69.4 & --    \\
UniTok~\cite{ma2025unitok} &7B&--&33.9&--&61.1&51.6&--&--&--&--&--\\
MetaMorph~\cite{tong2024metamorph} & 8B  & 71.8 & -- & -- & -- & 60.5 & -- & 58.3 & 41.8 & 75.8 & --   \\
\midrule
\multicolumn{12}{l}{\textcolor[RGB]{105, 105, 105}{\textit{Discrete Visual Input}}} \\
\midrule
Chameleon~\cite{team2024chameleon} & 34B  & -- & -- & 31.8 &   69.6 & -- & -- & -- & -- & -- & --   \\
SEED~\cite{li2023seed} &  13B   & 53.7 & -- & 33.1   & 44.8 & -- & -- & -- & -- & -- & -- \\
Show-o~\cite{xie2024show}  &  1.3B   & -- & -- & --  & 58.0 & -- & -- & 26.7 & -- & -- & 1097.2   \\
O-Mamba~\cite{zou2025omnimamba} & 1.3B & -- & -- & --&  60.8 & -- & --&  30.6& -- & --& -- \\
Harmon~\cite{wu2025harmonizing} & 1.5B & 67.1 & -- & -- & 58.9 & -- & -- & -- & 38.9 & 65.5 & 1476.0\\
VILA-U~\cite{wu2024vila}  &  7B   & 59.0 & 33.5 & --   & 60.8 & 60.8 & -- & -- & -- & -- & 1401.8   \\
EMU3~\cite{wang2024emu3}  & 8B   & 68.2 & 37.2 & 46.6 & 60.3 & 64.7 & 70.0 & 57.4 & 31.6 & 58.3 & 1509.9   \\
TokenFlow~\cite{qu2024tokenflow} &  14B  & 72.6 & 48.2 & --   & 62.5 & 62.3 & 75.8 & 56.6 & 43.2 & 76.8 & 1922.2  \\
UniToken~\cite{jiao2025unitoken} &  7B  & 70.3 & -- & 46.1   & -- & -- & 68.7 & -- & 34.2 & 71.1 & --  \\
MUSE-VL~\cite{xie2024muse} &  7B  & 69.1 & -- & 49.6   &--  &--  &69.8 &--  & 39.7 & 72.1 & --  \\
TokLIP~\cite{lin2025toklipmarryvisualtokens} &  7B  & 70.4 & 29.8 & --   & 59.5 &--  & -- &--  & 43.1 & 67.6 & --  \\
\rowcolor{gray!10} \textbf{UniCode²}~(Ours) &  7B & 73.5 & 48.6 & 54.0 &  69.2 & 65.7 & 79.6 & 65.4 & 47.1 & 79.0 & 2052.0  \\
\bottomrule
\end{tabular}
}
\caption{Evaluation on multimodal understanding benchmarks. 
}
\label{table_und}
\end{table*}

\setlength{\tabcolsep}{3pt}
\begin{table*}[t]
\centering
\small
\adjustbox{max width=\textwidth}{
\begin{tabular}{l | c|ccccc ccccc}
\toprule
\multirow{2}{*}{\textbf{Method}}  & \multirow{2}{*}{\textbf{Type}}  & \multicolumn{5}{c}{\textbf{GenEval}~$\uparrow$} & \multicolumn{5}{c}{\textbf{DPG-Bench}~$\uparrow$ } \\
\cmidrule(lr){3-7} \cmidrule(lr){8-12}
& & Single. & Two. & Count. & Colors & \textbf{Overall} & Global & Entity & Attribute & Other & \textbf{Overall} \\
\midrule
SD v2.1~\cite{rombach2022high} & \multirow{4}{*}{\textit{Diff.}}  & 0.98 &0.51 &0.44 &0.85 & 0.50 &-- &-- &-- &-- & -- \\
SDXL~\cite{podell2023sdxl} & & 0.98 & 0.74 &0.39 &0.85 &0.55 &83.3& 82.4 &80.9 &80.4&74.7  \\
PixArt-$\alpha$~\cite{chen2023pixart}  & & 0.98 & 0.50 & 0.44 & 0.80 & 0.48 & 75.0& 79.3& 78.6 &77.0 & 71.1 \\
DALL-E 3~\cite{betker2023improving}  & & 0.96 &0.87& 0.47& 0.83 &{0.67} &-- &-- &-- &-- & 83.5 \\
\midrule
Chameleon~\cite{team2024chameleon} & \multirow{5}{*}{\textit{AR.}} & -- & -- & -- & -- & 0.63 & -- & -- & -- & -- & -- \\
LlamaGen~\cite{sun2024autoregressive}  & & 0.71& 0.34&0.21 &0.58 &0.32 &-- &-- &-- &-- & --  \\
EMU3~\cite{wang2024emu3} & & 0.98 & 0.71 & 0.34 & 0.81  & 0.54 & 85.2 &86.7 &86.8 &90.2 & 80.1  \\
VAR~\cite{tian2024visual} & & -- & -- & -- & -- & 0.53 & -- & -- & -- & -- & 71.1  \\
TokenFlow~\cite{qu2024tokenflow} & & -- & -- & -- & -- & 0.63 & -- & -- & -- & -- & 73.4  \\

\midrule
Show-o~\cite{xie2024show}  & \multirow{6}{*}{\textit{AR. + Diff.}}  & 0.95 & 0.52 & 0.49 & 0.82 & 0.53 & --&-- &-- &-- & --  \\
Transfusion~\cite{zhou2024transfusion}  &  & -- & -- & -- & -- & 0.63 & -- & -- & -- & -- & -- \\
UniToken~\cite{ma2025unitok} & & 0.99 & 0.80 & 0.35 & 0.84 & 0.63 & -- & -- & --  & -- & --\\
ILLUME~\cite{wang2024illume} & & 0.99 & 0.86 & 0.45 & 0.71 & 0.61 & -- & -- & -- & -- & -- \\
JanusFlow~\cite{ma2024janusflow}  &  & 0.97 & 0.59 & 0.45 & 0.83 & 0.63 & 87.0 & 87.3 & 87.4 &88.1 &80.1\\
\rowcolor{gray!10} \textbf{UniCode²}~(Ours) & & 0.98 & 0.82 & 0.64 & 0.79 & {0.65} & 90.8 & 89.0 & 87.5 & 88.5 & {83.5} \\
\bottomrule
\end{tabular}
}
\caption{Visual generation performance comparison on GenEval~\cite{ghosh2023geneval} and DPG-Bench~\cite{hu2024ella}. The table below displays the first 4 metrics for each benchmark along with the overall score. \textit{Diff.} stands for \textit{Diffusion}, and \textit{AR.} stands for \textit{AutoRegressive}.}
\label{table_gen}
\end{table*}

\paragraph{Results on Generation Tasks.}
We evaluate UniCode² on GenEval and DPO-Bench, covering both short and compositional prompts. As shown in Table~\ref{table_gen}, UniCode² performs on par with the strongest diffusion-only, autoregressive-only, and hybrid approaches. 
UniCode² trains efficiently via autoregressive modeling with contrastive alignment, while keeping the diffusion decoder frozen. This design enables high-quality generation with a low cost, validating the strength of our semantically aligned, decoding-friendly token space.

\subsection{Ablation Studies}

\begin{table}[htb]
\centering
\small
\setlength{\tabcolsep}{8pt}
\begin{tabular}{lcccccc}
\toprule
\multirow{2}{*}{\textbf{Codebook Type}} & \multirow{2}{*}{\textbf{Codebook Size}} & \multicolumn{4}{c}{{\textbf{Performance}}} & \multirow{2}{*}{\textbf{Utilization} (\%)} \\
\cmidrule(lr){3-6}
& & SEEDB & AI2D & MMB &  \textbf{Average} & \\
\midrule
\multirow{3}{*}{\makecell{{VQGAN~\cite{esser2021taming}} \\ (\textit{pixel recon.})}}
& 10K & 63.1 & 56.2 & 63.1 & 64.1 & 98.5 \\
& 100K & 51.4 & 56.1 & 54.7 & 54.1 & 12.4 \\
& 500K & 46.0 & 49.8 & 48.8 & 48.2 & 2.3 \\
\midrule
\multirow{3}{*}{\textbf{Ours} (\textit{clustering})} 
 & 10K & 69.9 & 76.8 & 77.1 & 74.6 & 99.5 \\
 & 100K & 72.4 & 78.8 & 77.6 & 76.3 & 99.1 \\
\rowcolor{gray!10}  & 500K & 73.5 & 79.6 & 79.0  & 77.4  & 98.7 \\
\bottomrule
\end{tabular}
\caption{Codebook size ablation: impact on performance and utilization.}
\label{table_size}
\end{table}

\paragraph{Effect of Codebook Size.}
We evaluate codebook sizes (10K, 100K, 500K) using two approaches: the VQGAN codebook~\cite{esser2021taming} commonly used in Unified MLLMs, trained on pixel reconstruction tasks and lacking inherent semantic alignment, and our semantic-based codebook, constructed by clustering SigLIP embeddings to ensure natural semantic alignment and high utilization. As shown in Table~\ref{table_size}, VQGAN-based codebooks suffer severe degradation with larger vocabularies—performance drops and utilization collapses to 2.3\% at 500K due to inactive tokens. In contrast, our semantic-based codebook achieves consistently high utilization (\textasciitilde99\%) across all sizes, with larger vocabularies improving benchmark performance. Utilization is computed as the fraction of codewords activated at least once over the validation set. These results demonstrate that effective scaling requires well-aligned semantics and diverse token usage, both enabled by our semantic-based clustering approach.

\paragraph{Effect of Cascaded Codebook.}
Table~\ref{table_cascaded} compares codebook configurations and direct visual encoder training. Our cascaded design achieves the best understanding performance by decoupling indexing (frozen codebook) and refinement (trainable codebook), ensuring stable training and robust semantic representation. In contrast, a single trainable codebook leads to optimization instability and degraded performance, highlighting the importance of combining frozen mapping with adaptive refinement. Additionally, our approach outperforms direct encoder training with 40\% fewer parameters, demonstrating its efficiency and effectiveness.

\begin{table*}[htb]
\begin{floatrow}
\capbtabbox{
\small
\begin{tabular}{lcc}
\toprule
\textbf{Setting} & \textbf{\# Params} & \textbf{Avg.} \\
\midrule
Train Encoder (No Codebook)      & 3.3~GB & 75.6 \\
\midrule
One Codebook (Frozen) & 0~~~~GB  & 72.2 \\
One Codebook (Trainable) & 2.0 GB  & 69.7 \\
Both Codebooks (Trainable) & 4.0~GB & 67.1 \\
\rowcolor{gray!10}  {\textbf{Cascaded Codebook} (Ours)} & 2.0 GB  & {77.4} \\
\bottomrule
\end{tabular}
}{
 \caption{Ablation on cascaded codebook: \# Params denotes \# trainable parameters, Avg. is mean result across SEEDB, AI2D, and MMB.}
 \label{table_cascaded}
}
\capbtabbox{
\small
\begin{tabular}{lcccc}
\toprule
\textbf{Setting} & \textbf{Unet-\textit{ft}?} & \textbf{\# Data} &  \textbf{GenEval} & \textbf{DPG} \\
\midrule
SDXL~\cite{podell2023sdxl} & \checkmark    & --           & 0.55 & 74.7 \\
\midrule
\rowcolor{gray!10}\textbf{Ours} & \ding{53} & 200K    & 0.62 & 77.3 \\
Unet-\textit{ft}   & \checkmark & 200K          & 0.61 & 78.4 \\
20K Data  & \ding{53}   & 20K       & 0.59 & 76.5 \\
Full Data & \ding{53} & 4M & 0.62& 77.1 \\
\bottomrule
\end{tabular}
}{
 \caption{Generation quality comparison across settings, evaluated on GenEval and DPG-Bench Overall metrics. \# Unet-ft means Unet-finetune.}
 \label{table_plugin}
}
\end{floatrow}
\end{table*}

\paragraph{Plug-in Generation Quality.}
Beyond the results in Table~\ref{table_gen} with FLUX~\cite{blackforestlabs2024}, we showcase seamless integration with SDXL~\cite{podell2023sdxl} in Table~\ref{table_plugin}. Using 200K JourneyDB image-text pairs for contrastive learning, our approach improves SDXL performance, highlighting the semantic alignment and high-capacity representation enabled by the codebook. Fine-tuning the Unet further refines task-specific representations, boosting performance. Notably, with just 20K pairs, our method achieves competitive results, demonstrating data efficiency. Scaling to the full dataset slightly reduces performance compared to 200K pairs, suggesting diminishing returns or noise in overly large datasets.


\begin{figure*}[t!]
  \centering
  \includegraphics[width=1.0\linewidth]{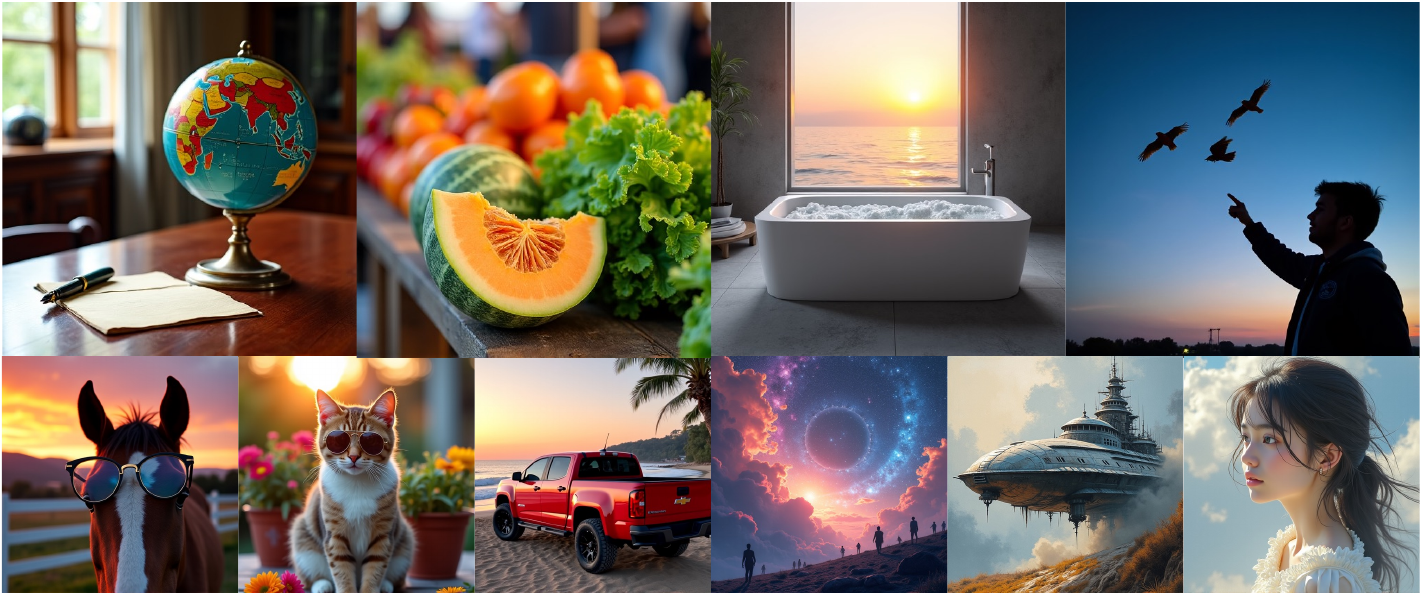}
  \caption{Images generated in a resolution of $512 \times 512$ with UniCode².}
  \label{fig_gen_imgs}
\end{figure*}

\paragraph{Generated Image Samples.} Figure~\ref{fig_gen_imgs} showcases images generated by UniCode² from text prompts sampled from DPG-bench. Benefiting from its semantically grounded codebook, our model demonstrates strong generation capabilities not only on richly detailed real-world scenes (top row) but also on imaginative and counterintuitive concepts (bottom row).


\section{Conclusion}


We introduce UniCode², a cascaded codebook framework that overcomes key limitations of existing unified multimodal large language models. By clustering SigLIP embeddings into a 500K-entry codebook, UniCode² achieves rich semantic alignment and expanded visual token capacity. Its decoupled design—combining a frozen codebook for stable indexing with a trainable codebook for task-specific refinement—ensures stable training and high token utilization. This modular architecture seamlessly integrates with pretrained diffusion decoders, enabling efficient, high-quality text-to-image generation. Experiments show UniCode² scales visual token spaces effectively while maintaining stability, semantic fidelity, and modularity, advancing unified multimodal modeling.

\paragraph{Limitations.} 
Our approach employs a fixed codebook size, which, while enabling stable and efficient tokenization, may restrict adaptability to diverse data distributions or evolving task demands. Dynamically adjusting codebook capacity remains an open direction for future work.

\bibliography{A-reference.bib}

\begin{thebibliography}{10}

\bibitem{achiam2023gpt}
Josh Achiam, Steven Adler, Sandhini Agarwal, Lama Ahmad, Ilge Akkaya, Florencia~Leoni Aleman, Diogo Almeida, Janko Altenschmidt, Sam Altman, Shyamal Anadkat, et~al.
\newblock Gpt-4 technical report.
\newblock {\em arXiv preprint arXiv:2303.08774}, 2023.

\bibitem{awais2025foundation}
Muhammad Awais, Muzammal Naseer, Salman Khan, Rao~Muhammad Anwer, Hisham Cholakkal, Mubarak Shah, Ming-Hsuan Yang, and Fahad~Shahbaz Khan.
\newblock Foundation models defining a new era in vision: a survey and outlook.
\newblock {\em IEEE Transactions on Pattern Analysis and Machine Intelligence}, 2025.

\bibitem{bai2023qwen}
Jinze Bai, Shuai Bai, Yunfei Chu, Zeyu Cui, Kai Dang, Xiaodong Deng, Yang Fan, Wenbin Ge, Yu~Han, Fei Huang, et~al.
\newblock Qwen technical report.
\newblock {\em arXiv preprint arXiv:2309.16609}, 2023.

\bibitem{bai2025qwen2}
Shuai Bai, Keqin Chen, Xuejing Liu, Jialin Wang, Wenbin Ge, Sibo Song, Kai Dang, Peng Wang, Shijie Wang, Jun Tang, et~al.
\newblock Qwen2. 5-vl technical report.
\newblock {\em arXiv preprint arXiv:2502.13923}, 2025.

\bibitem{bai2024factorized}
Zechen Bai, Jianxiong Gao, Ziteng Gao, Pichao Wang, Zheng Zhang, Tong He, and Mike~Zheng Shou.
\newblock Factorized visual tokenization and generation.
\newblock {\em arXiv preprint arXiv:2411.16681}, 2024.

\bibitem{bao2021beit}
Hangbo Bao, Li~Dong, Songhao Piao, and Furu Wei.
\newblock Beit: Bert pre-training of image transformers.
\newblock {\em arXiv preprint arXiv:2106.08254}, 2021.

\bibitem{betker2023improving}
James Betker, Gabriel Goh, Li~Jing, Tim Brooks, Jianfeng Wang, Linjie Li, Long Ouyang, Juntang Zhuang, Joyce Lee, Yufei Guo, et~al.
\newblock Improving image generation with better captions.
\newblock {\em Computer Science. https://cdn. openai. com/papers/dall-e-3. pdf}, 2(3):8, 2023.

\bibitem{cao2023efficient}
Shiyue Cao, Yueqin Yin, Lianghua Huang, Yu~Liu, Xin Zhao, Deli Zhao, and Kaigi Huang.
\newblock Efficient-vqgan: Towards high-resolution image generation with efficient vision transformers.
\newblock In {\em Proceedings of the IEEE/CVF International Conference on Computer Vision}, pages 7368--7377, 2023.

\bibitem{chen2023pixart}
Junsong Chen, Jincheng Yu, Chongjian Ge, Lewei Yao, Enze Xie, Yue Wu, Zhongdao Wang, James Kwok, Ping Luo, Huchuan Lu, et~al.
\newblock Pixart-alpha: Fast training of diffusion transformer for photorealistic text-to-image synthesis.
\newblock {\em arXiv preprint arXiv:2310.00426}, 2023.

\bibitem{chen2024sharegpt4v}
Lin Chen, Jinsong Li, Xiaoyi Dong, Pan Zhang, Conghui He, Jiaqi Wang, Feng Zhao, and Dahua Lin.
\newblock Sharegpt4v: Improving large multi-modal models with better captions.
\newblock In {\em European Conference on Computer Vision}, pages 370--387. Springer, 2024.

\bibitem{chen2024we}
Lin Chen, Jinsong Li, Xiaoyi Dong, Pan Zhang, Yuhang Zang, Zehui Chen, Haodong Duan, Jiaqi Wang, Yu~Qiao, Dahua Lin, et~al.
\newblock Are we on the right way for evaluating large vision-language models?
\newblock {\em arXiv preprint arXiv:2403.20330}, 2024.

\bibitem{chenmai}
Yanzhe Chen, Zhiwen Yang, Jinglin Xu, and Yuxin Peng.
\newblock Mai: A multi-turn aggregation-iteration model for composed image retrieval.
\newblock In {\em The Thirteenth International Conference on Learning Representations}, 2025.

\bibitem{chen2024expanding}
Zhe Chen, Weiyun Wang, Yue Cao, Yangzhou Liu, Zhangwei Gao, Erfei Cui, Jinguo Zhu, Shenglong Ye, Hao Tian, Zhaoyang Liu, et~al.
\newblock Expanding performance boundaries of open-source multimodal models with model, data, and test-time scaling.
\newblock {\em arXiv preprint arXiv:2412.05271}, 2024.

\bibitem{chen2025semhitok}
Zisheng Chen, Chunwei Wang, Xiuwei Chen, Hang Xu, Jianhua Han, and Xiaodan Liang.
\newblock Semhitok: A unified image tokenizer via semantic-guided hierarchical codebook for multimodal understanding and generation.
\newblock {\em arXiv preprint arXiv:2503.06764}, 2025.

\bibitem{chern2024anole}
Ethan Chern, Jiadi Su, Yan Ma, and Pengfei Liu.
\newblock Anole: An open, autoregressive, native large multimodal models for interleaved image-text generation.
\newblock {\em arXiv preprint arXiv:2407.06135}, 2024.

\bibitem{dong2023dreamllm}
Runpei Dong, Chunrui Han, Yuang Peng, Zekun Qi, Zheng Ge, Jinrong Yang, Liang Zhao, Jianjian Sun, Hongyu Zhou, Haoran Wei, et~al.
\newblock Dreamllm: Synergistic multimodal comprehension and creation.
\newblock {\em arXiv preprint arXiv:2309.11499}, 2023.

\bibitem{esser2021taming}
Patrick Esser, Robin Rombach, and Bjorn Ommer.
\newblock Taming transformers for high-resolution image synthesis.
\newblock In {\em Proceedings of the IEEE/CVF conference on computer vision and pattern recognition}, pages 12873--12883, 2021.

\bibitem{fang2024eva}
Yuxin Fang, Quan Sun, Xinggang Wang, Tiejun Huang, Xinlong Wang, and Yue Cao.
\newblock Eva-02: A visual representation for neon genesis.
\newblock {\em Image and Vision Computing}, 149:105171, 2024.

\bibitem{fang2023eva}
Yuxin Fang, Wen Wang, Binhui Xie, Quan Sun, Ledell Wu, Xinggang Wang, Tiejun Huang, Xinlong Wang, and Yue Cao.
\newblock Eva: Exploring the limits of masked visual representation learning at scale.
\newblock In {\em Proceedings of the IEEE/CVF conference on computer vision and pattern recognition}, pages 19358--19369, 2023.

\bibitem{ge2024seed}
Yuying Ge, Sijie Zhao, Jinguo Zhu, Yixiao Ge, Kun Yi, Lin Song, Chen Li, Xiaohan Ding, and Ying Shan.
\newblock Seed-x: Multimodal models with unified multi-granularity comprehension and generation.
\newblock {\em arXiv preprint arXiv:2404.14396}, 2024.

\bibitem{ghosh2023geneval}
Dhruba Ghosh, Hannaneh Hajishirzi, and Ludwig Schmidt.
\newblock Geneval: An object-focused framework for evaluating text-to-image alignment.
\newblock {\em Advances in Neural Information Processing Systems}, 36:52132--52152, 2023.

\bibitem{gui2024survey}
Jie Gui, Tuo Chen, Jing Zhang, Qiong Cao, Zhenan Sun, Hao Luo, and Dacheng Tao.
\newblock A survey on self-supervised learning: Algorithms, applications, and future trends.
\newblock {\em IEEE Transactions on Pattern Analysis and Machine Intelligence}, 2024.

\bibitem{hu2024ella}
Xiwei Hu, Rui Wang, Yixiao Fang, Bin Fu, Pei Cheng, and Gang Yu.
\newblock Ella: Equip diffusion models with llm for enhanced semantic alignment.
\newblock {\em arXiv preprint arXiv:2403.05135}, 2024.

\bibitem{huang2025illume}
Runhui Huang, Chunwei Wang, Junwei Yang, Guansong Lu, Yunlong Yuan, Jianhua Han, Lu~Hou, Wei Zhang, Lanqing Hong, Hengshuang Zhao, et~al.
\newblock Illume+: Illuminating unified mllm with dual visual tokenization and diffusion refinement.
\newblock {\em arXiv preprint arXiv:2504.01934}, 2025.

\bibitem{hudson2019gqa}
Drew~A Hudson and Christopher~D Manning.
\newblock Gqa: A new dataset for real-world visual reasoning and compositional question answering.
\newblock In {\em Proceedings of the IEEE/CVF conference on computer vision and pattern recognition}, pages 6700--6709, 2019.

\bibitem{jiao2025unitoken}
Yang Jiao, Haibo Qiu, Zequn Jie, Shaoxiang Chen, Jingjing Chen, Lin Ma, and Yu-Gang Jiang.
\newblock Unitoken: Harmonizing multimodal understanding and generation through unified visual encoding.
\newblock {\em arXiv preprint arXiv:2504.04423}, 2025.

\bibitem{kembhavi2016diagram}
Aniruddha Kembhavi, Mike Salvato, Eric Kolve, Minjoon Seo, Hannaneh Hajishirzi, and Ali Farhadi.
\newblock A diagram is worth a dozen images.
\newblock In {\em Computer Vision--ECCV 2016: 14th European Conference, Amsterdam, The Netherlands, October 11--14, 2016, Proceedings, Part IV 14}, pages 235--251. Springer, 2016.

\bibitem{blackforestlabs2024}
Black~Forest Labs.
\newblock \url{https://github.com/black-forest-labs/flux}, 2 2024.

\bibitem{lee2022autoregressive}
Doyup Lee, Chiheon Kim, Saehoon Kim, Minsu Cho, and Wook-Shin Han.
\newblock Autoregressive image generation using residual quantization.
\newblock In {\em Proceedings of the IEEE/CVF Conference on Computer Vision and Pattern Recognition}, pages 11523--11532, 2022.

\bibitem{li2024llava}
Bo~Li, Yuanhan Zhang, Dong Guo, Renrui Zhang, Feng Li, Hao Zhang, Kaichen Zhang, Peiyuan Zhang, Yanwei Li, Ziwei Liu, et~al.
\newblock Llava-onevision: Easy visual task transfer.
\newblock {\em arXiv preprint arXiv:2408.03326}, 2024.

\bibitem{li2023seed}
Bohao Li, Rui Wang, Guangzhi Wang, Yuying Ge, Yixiao Ge, and Ying Shan.
\newblock Seed-bench: Benchmarking multimodal llms with generative comprehension.
\newblock {\em arXiv preprint arXiv:2307.16125}, 2023.

\bibitem{liang2024survey}
Zijing Liang, Yanjie Xu, Yifan Hong, Penghui Shang, Qi~Wang, Qiang Fu, and Ke~Liu.
\newblock A survey of multimodel large language models.
\newblock In {\em Proceedings of the 3rd International Conference on Computer, Artificial Intelligence and Control Engineering}, pages 405--409, 2024.

\bibitem{lin2025toklipmarryvisualtokens}
Haokun Lin, Teng Wang, Yixiao Ge, Yuying Ge, Zhichao Lu, Ying Wei, Qingfu Zhang, Zhenan Sun, and Ying Shan.
\newblock Toklip: Marry visual tokens to clip for multimodal comprehension and generation, 2025.

\bibitem{liu2024world}
Hao Liu, Wilson Yan, Matei Zaharia, and Pieter Abbeel.
\newblock World model on million-length video and language with blockwise ringattention.
\newblock {\em arXiv preprint arXiv:2402.08268}, 2024.

\bibitem{liu2024improved}
Haotian Liu, Chunyuan Li, Yuheng Li, and Yong~Jae Lee.
\newblock Improved baselines with visual instruction tuning.
\newblock In {\em Proceedings of the IEEE/CVF Conference on Computer Vision and Pattern Recognition}, pages 26296--26306, 2024.

\bibitem{liu2023visual}
Haotian Liu, Chunyuan Li, Qingyang Wu, and Yong~Jae Lee.
\newblock Visual instruction tuning.
\newblock {\em Advances in neural information processing systems}, 36:34892--34916, 2023.

\bibitem{liu2024mmbench}
Yuan Liu, Haodong Duan, Yuanhan Zhang, Bo~Li, Songyang Zhang, Wangbo Zhao, Yike Yuan, Jiaqi Wang, Conghui He, Ziwei Liu, et~al.
\newblock Mmbench: Is your multi-modal model an all-around player?
\newblock In {\em European conference on computer vision}, pages 216--233. Springer, 2024.

\bibitem{loshchilov2017decoupled}
Ilya Loshchilov and Frank Hutter.
\newblock Decoupled weight decay regularization.
\newblock {\em arXiv preprint arXiv:1711.05101}, 2017.

\bibitem{lu2024ovis}
Shiyin Lu, Yang Li, Qing-Guo Chen, Zhao Xu, Weihua Luo, Kaifu Zhang, and Han-Jia Ye.
\newblock Ovis: Structural embedding alignment for multimodal large language model.
\newblock {\em arXiv preprint arXiv:2405.20797}, 2024.

\bibitem{ma2025unitok}
Chuofan Ma, Yi~Jiang, Junfeng Wu, Jihan Yang, Xin Yu, Zehuan Yuan, Bingyue Peng, and Xiaojuan Qi.
\newblock Unitok: A unified tokenizer for visual generation and understanding.
\newblock {\em arXiv preprint arXiv:2502.20321}, 2025.

\bibitem{ma2024janusflow}
Yiyang Ma, Xingchao Liu, Xiaokang Chen, Wen Liu, Chengyue Wu, Zhiyu Wu, Zizheng Pan, Zhenda Xie, Haowei Zhang, Liang Zhao, et~al.
\newblock Janusflow: Harmonizing autoregression and rectified flow for unified multimodal understanding and generation.
\newblock {\em arXiv preprint arXiv:2411.07975}, 2024.

\bibitem{peng2022beit}
Zhiliang Peng, Li~Dong, Hangbo Bao, Qixiang Ye, and Furu Wei.
\newblock Beit v2: Masked image modeling with vector-quantized visual tokenizers.
\newblock {\em arXiv preprint arXiv:2208.06366}, 2022.

\bibitem{podell2023sdxl}
Dustin Podell, Zion English, Kyle Lacey, Andreas Blattmann, Tim Dockhorn, Jonas M{\"u}ller, Joe Penna, and Robin Rombach.
\newblock Sdxl: Improving latent diffusion models for high-resolution image synthesis.
\newblock {\em arXiv preprint arXiv:2307.01952}, 2023.

\bibitem{qu2024tokenflow}
Liao Qu, Huichao Zhang, Yiheng Liu, Xu~Wang, Yi~Jiang, Yiming Gao, Hu~Ye, Daniel~K Du, Zehuan Yuan, and Xinglong Wu.
\newblock Tokenflow: Unified image tokenizer for multimodal understanding and generation.
\newblock {\em arXiv preprint arXiv:2412.03069}, 2024.

\bibitem{ren2025beyond}
Sucheng Ren, Qihang Yu, Ju~He, Xiaohui Shen, Alan Yuille, and Liang-Chieh Chen.
\newblock Beyond next-token: Next-x prediction for autoregressive visual generation.
\newblock {\em arXiv preprint arXiv:2502.20388}, 2025.

\bibitem{rombach2022high}
Robin Rombach, Andreas Blattmann, Dominik Lorenz, Patrick Esser, and Bj{\"o}rn Ommer.
\newblock High-resolution image synthesis with latent diffusion models.
\newblock In {\em Proceedings of the IEEE/CVF conference on computer vision and pattern recognition}, pages 10684--10695, 2022.

\bibitem{shi2024taming}
Fengyuan Shi, Zhuoyan Luo, Yixiao Ge, Yujiu Yang, Ying Shan, and Limin Wang.
\newblock Taming scalable visual tokenizer for autoregressive image generation.
\newblock {\em arXiv preprint arXiv:2412.02692}, 2024.

\bibitem{singh2019towards}
Amanpreet Singh, Vivek Natarajan, Meet Shah, Yu~Jiang, Xinlei Chen, Dhruv Batra, Devi Parikh, and Marcus Rohrbach.
\newblock Towards vqa models that can read.
\newblock In {\em Proceedings of the IEEE/CVF conference on computer vision and pattern recognition}, pages 8317--8326, 2019.

\bibitem{song2025dualtoken}
Wei Song, Yuran Wang, Zijia Song, Yadong Li, Haoze Sun, Weipeng Chen, Zenan Zhou, Jianhua Xu, Jiaqi Wang, and Kaicheng Yu.
\newblock Dualtoken: Towards unifying visual understanding and generation with dual visual vocabularies.
\newblock {\em arXiv preprint arXiv:2503.14324}, 2025.

\bibitem{sun2023journeydb}
Keqiang Sun, Junting Pan, Yuying Ge, Hao Li, Haodong Duan, Xiaoshi Wu, Renrui Zhang, Aojun Zhou, Zipeng Qin, Yi~Wang, et~al.
\newblock Journeydb: A benchmark for generative image understanding.
\newblock {\em Advances in neural information processing systems}, 36:49659--49678, 2023.

\bibitem{sun2024autoregressive}
Peize Sun, Yi~Jiang, Shoufa Chen, Shilong Zhang, Bingyue Peng, Ping Luo, and Zehuan Yuan.
\newblock Autoregressive model beats diffusion: Llama for scalable image generation.
\newblock {\em arXiv preprint arXiv:2406.06525}, 2024.

\bibitem{sun2024generative}
Quan Sun, Yufeng Cui, Xiaosong Zhang, Fan Zhang, Qiying Yu, Yueze Wang, Yongming Rao, Jingjing Liu, Tiejun Huang, and Xinlong Wang.
\newblock Generative multimodal models are in-context learners.
\newblock In {\em Proceedings of the IEEE/CVF Conference on Computer Vision and Pattern Recognition}, pages 14398--14409, 2024.

\bibitem{sun2023emu}
Quan Sun, Qiying Yu, Yufeng Cui, Fan Zhang, Xiaosong Zhang, Yueze Wang, Hongcheng Gao, Jingjing Liu, Tiejun Huang, and Xinlong Wang.
\newblock Emu: Generative pretraining in multimodality.
\newblock {\em arXiv preprint arXiv:2307.05222}, 2023.

\bibitem{team2024chameleon}
Chameleon Team.
\newblock Chameleon: Mixed-modal early-fusion foundation models.
\newblock {\em arXiv preprint arXiv:2405.09818}, 2024.

\bibitem{team2023gemini}
Gemini Team, Rohan Anil, Sebastian Borgeaud, Jean-Baptiste Alayrac, Jiahui Yu, Radu Soricut, Johan Schalkwyk, Andrew~M Dai, Anja Hauth, Katie Millican, et~al.
\newblock Gemini: a family of highly capable multimodal models.
\newblock {\em arXiv preprint arXiv:2312.11805}, 2023.

\bibitem{tian2024visual}
Keyu Tian, Yi~Jiang, Zehuan Yuan, Bingyue Peng, and Liwei Wang.
\newblock Visual autoregressive modeling: Scalable image generation via next-scale prediction.
\newblock {\em Advances in neural information processing systems}, 37:84839--84865, 2024.

\bibitem{tong2024metamorph}
Shengbang Tong, David Fan, Jiachen Zhu, Yunyang Xiong, Xinlei Chen, Koustuv Sinha, Michael Rabbat, Yann LeCun, Saining Xie, and Zhuang Liu.
\newblock Metamorph: Multimodal understanding and generation via instruction tuning.
\newblock {\em arXiv preprint arXiv:2412.14164}, 2024.

\bibitem{van2017neural}
Aaron Van Den~Oord, Oriol Vinyals, et~al.
\newblock Neural discrete representation learning.
\newblock {\em Advances in neural information processing systems}, 30, 2017.

\bibitem{wang2025discrete}
Bohan Wang, Zhongqi Yue, Fengda Zhang, Shuo Chen, Li'an Bi, Junzhe Zhang, Xue Song, Kennard~Yanting Chan, Jiachun Pan, Weijia Wu, et~al.
\newblock Discrete visual tokens of autoregression, by diffusion, and for reasoning.
\newblock {\em arXiv preprint arXiv:2505.07538}, 2025.

\bibitem{wang2024illume}
Chunwei Wang, Guansong Lu, Junwei Yang, Runhui Huang, Jianhua Han, Lu~Hou, Wei Zhang, and Hang Xu.
\newblock Illume: Illuminating your llms to see, draw, and self-enhance.
\newblock {\em arXiv preprint arXiv:2412.06673}, 2024.

\bibitem{wang2024omnitokenizer}
Junke Wang, Yi~Jiang, Zehuan Yuan, Bingyue Peng, Zuxuan Wu, and Yu-Gang Jiang.
\newblock Omnitokenizer: A joint image-video tokenizer for visual generation.
\newblock {\em Advances in Neural Information Processing Systems}, 37:28281--28295, 2024.

\bibitem{wang2024image}
Luting Wang, Yang Zhao, Zijian Zhang, Jiashi Feng, Si~Liu, and Bingyi Kang.
\newblock Image understanding makes for a good tokenizer for image generation.
\newblock {\em Advances in Neural Information Processing Systems}, 37:31015--31035, 2024.

\bibitem{wang2024qwen2}
Peng Wang, Shuai Bai, Sinan Tan, Shijie Wang, Zhihao Fan, Jinze Bai, Keqin Chen, Xuejing Liu, Jialin Wang, Wenbin Ge, et~al.
\newblock Qwen2-vl: Enhancing vision-language model's perception of the world at any resolution.
\newblock {\em arXiv preprint arXiv:2409.12191}, 2024.

\bibitem{wang2023image}
Wenhui Wang, Hangbo Bao, Li~Dong, Johan Bjorck, Zhiliang Peng, Qiang Liu, Kriti Aggarwal, Owais~Khan Mohammed, Saksham Singhal, Subhojit Som, et~al.
\newblock Image as a foreign language: Beit pretraining for vision and vision-language tasks.
\newblock In {\em Proceedings of the IEEE/CVF Conference on Computer Vision and Pattern Recognition}, pages 19175--19186, 2023.

\bibitem{wang2024emu3}
Xinlong Wang, Xiaosong Zhang, Zhengxiong Luo, Quan Sun, Yufeng Cui, Jinsheng Wang, Fan Zhang, Yueze Wang, Zhen Li, Qiying Yu, et~al.
\newblock Emu3: Next-token prediction is all you need.
\newblock {\em arXiv preprint arXiv:2409.18869}, 2024.

\bibitem{wu2024janus}
Chengyue Wu, Xiaokang Chen, Zhiyu Wu, Yiyang Ma, Xingchao Liu, Zizheng Pan, Wen Liu, Zhenda Xie, Xingkai Yu, Chong Ruan, et~al.
\newblock Janus: Decoupling visual encoding for unified multimodal understanding and generation.
\newblock {\em arXiv preprint arXiv:2410.13848}, 2024.

\bibitem{wu2024liquid}
Junfeng Wu, Yi~Jiang, Chuofan Ma, Yuliang Liu, Hengshuang Zhao, Zehuan Yuan, Song Bai, and Xiang Bai.
\newblock Liquid: Language models are scalable multi-modal generators.
\newblock {\em arXiv preprint arXiv:2412.04332}, 2024.

\bibitem{wu2024next}
Shengqiong Wu, Hao Fei, Leigang Qu, Wei Ji, and Tat-Seng Chua.
\newblock Next-gpt: Any-to-any multimodal llm.
\newblock In {\em Forty-first International Conference on Machine Learning}, 2024.

\bibitem{wu2025harmonizing}
Size Wu, Wenwei Zhang, Lumin Xu, Sheng Jin, Zhonghua Wu, Qingyi Tao, Wentao Liu, Wei Li, and Chen~Change Loy.
\newblock Harmonizing visual representations for unified multimodal understanding and generation.
\newblock {\em arXiv preprint arXiv:2503.21979}, 2025.

\bibitem{wu2024vila}
Yecheng Wu, Zhuoyang Zhang, Junyu Chen, Haotian Tang, Dacheng Li, Yunhao Fang, Ligeng Zhu, Enze Xie, Hongxu Yin, Li~Yi, et~al.
\newblock Vila-u: a unified foundation model integrating visual understanding and generation.
\newblock {\em arXiv preprint arXiv:2409.04429}, 2024.

\bibitem{realworldqa}
x.ai.
\newblock Grok-1.5 vision preview, 6 2024.

\bibitem{xiao2024omnigen}
Shitao Xiao, Yueze Wang, Junjie Zhou, Huaying Yuan, Xingrun Xing, Ruiran Yan, Chaofan Li, Shuting Wang, Tiejun Huang, and Zheng Liu.
\newblock Omnigen: Unified image generation.
\newblock {\em arXiv preprint arXiv:2409.11340}, 2024.

\bibitem{xie2024show}
Jinheng Xie, Weijia Mao, Zechen Bai, David~Junhao Zhang, Weihao Wang, Kevin~Qinghong Lin, Yuchao Gu, Zhijie Chen, Zhenheng Yang, and Mike~Zheng Shou.
\newblock Show-o: One single transformer to unify multimodal understanding and generation.
\newblock {\em arXiv preprint arXiv:2408.12528}, 2024.

\bibitem{xie2024muse}
Rongchang Xie, Chen Du, Ping Song, and Chang Liu.
\newblock Muse-vl: Modeling unified vlm through semantic discrete encoding.
\newblock {\em arXiv preprint arXiv:2411.17762}, 2024.

\bibitem{xiong2024efficientsam}
Yunyang Xiong, Bala Varadarajan, Lemeng Wu, Xiaoyu Xiang, Fanyi Xiao, Chenchen Zhu, Xiaoliang Dai, Dilin Wang, Fei Sun, Forrest Iandola, et~al.
\newblock Efficientsam: Leveraged masked image pretraining for efficient segment anything.
\newblock In {\em Proceedings of the IEEE/CVF Conference on Computer Vision and Pattern Recognition}, pages 16111--16121, 2024.

\bibitem{yang2024qwen2}
An~Yang, Baosong Yang, Beichen Zhang, Binyuan Hui, Bo~Zheng, Bowen Yu, Chengyuan Li, Dayiheng Liu, Fei Huang, Haoran Wei, et~al.
\newblock Qwen2. 5 technical report.
\newblock {\em arXiv preprint arXiv:2412.15115}, 2024.

\bibitem{yu2021vector}
Jiahui Yu, Xin Li, Jing~Yu Koh, Han Zhang, Ruoming Pang, James Qin, Alexander Ku, Yuanzhong Xu, Jason Baldridge, and Yonghui Wu.
\newblock Vector-quantized image modeling with improved vqgan.
\newblock {\em arXiv preprint arXiv:2110.04627}, 2021.

\bibitem{yu2023language}
Lijun Yu, Jos{\'e} Lezama, Nitesh~B Gundavarapu, Luca Versari, Kihyuk Sohn, David Minnen, Yong Cheng, Vighnesh Birodkar, Agrim Gupta, Xiuye Gu, et~al.
\newblock Language model beats diffusion--tokenizer is key to visual generation.
\newblock {\em arXiv preprint arXiv:2310.05737}, 2023.

\bibitem{yu2023mm}
Weihao Yu, Zhengyuan Yang, Linjie Li, Jianfeng Wang, Kevin Lin, Zicheng Liu, Xinchao Wang, and Lijuan Wang.
\newblock Mm-vet: Evaluating large multimodal models for integrated capabilities.
\newblock {\em arXiv preprint arXiv:2308.02490}, 2023.

\bibitem{yue2024mmmu}
Xiang Yue, Yuansheng Ni, Kai Zhang, Tianyu Zheng, Ruoqi Liu, Ge~Zhang, Samuel Stevens, Dongfu Jiang, Weiming Ren, Yuxuan Sun, et~al.
\newblock Mmmu: A massive multi-discipline multimodal understanding and reasoning benchmark for expert agi.
\newblock In {\em Proceedings of the IEEE/CVF Conference on Computer Vision and Pattern Recognition}, pages 9556--9567, 2024.

\bibitem{zhai2023sigmoid}
Xiaohua Zhai, Basil Mustafa, Alexander Kolesnikov, and Lucas Beyer.
\newblock Sigmoid loss for language image pre-training.
\newblock In {\em Proceedings of the IEEE/CVF international conference on computer vision}, pages 11975--11986, 2023.

\bibitem{zhang2025token}
Haichao Zhang and Yun Fu.
\newblock Token dynamics: Towards efficient and dynamic video token representation for video large language models.
\newblock {\em arXiv preprint arXiv:2503.16980}, 2025.

\bibitem{zhang2025unified}
Xinjie Zhang, Jintao Guo, Shanshan Zhao, Minghao Fu, Lunhao Duan, Guo-Hua Wang, Qing-Guo Chen, Zhao Xu, Weihua Luo, and Kaifu Zhang.
\newblock Unified multimodal understanding and generation models: Advances, challenges, and opportunities.
\newblock {\em arXiv preprint arXiv:2505.02567}, 2025.

\bibitem{zhao2025qlip}
Yue Zhao, Fuzhao Xue, Scott Reed, Linxi Fan, Yuke Zhu, Jan Kautz, Zhiding Yu, Philipp Kr{\"a}henb{\"u}hl, and De-An Huang.
\newblock Qlip: Text-aligned visual tokenization unifies auto-regressive multimodal understanding and generation.
\newblock {\em arXiv preprint arXiv:2502.05178}, 2025.

\bibitem{zheng2023online}
Chuanxia Zheng and Andrea Vedaldi.
\newblock Online clustered codebook.
\newblock In {\em Proceedings of the IEEE/CVF International Conference on Computer Vision}, pages 22798--22807, 2023.

\bibitem{zhou2024transfusion}
Chunting Zhou, Lili Yu, Arun Babu, Kushal Tirumala, Michihiro Yasunaga, Leonid Shamis, Jacob Kahn, Xuezhe Ma, Luke Zettlemoyer, and Omer Levy.
\newblock Transfusion: Predict the next token and diffuse images with one multi-modal model.
\newblock {\em arXiv preprint arXiv:2408.11039}, 2024.

\bibitem{zhu2024scaling}
Lei Zhu, Fangyun Wei, Yanye Lu, and Dong Chen.
\newblock Scaling the codebook size of vqgan to 100,000 with a utilization rate of 99\%.
\newblock {\em arXiv preprint arXiv:2406.11837}, 2024.

\bibitem{zhu2024addressing}
Yongxin Zhu, Bocheng Li, Yifei Xin, and Linli Xu.
\newblock Addressing representation collapse in vector quantized models with one linear layer.
\newblock {\em arXiv preprint arXiv:2411.02038}, 2024.

\bibitem{zou2025omnimamba}
Jialv Zou, Bencheng Liao, Qian Zhang, Wenyu Liu, and Xinggang Wang.
\newblock Omnimamba: Efficient and unified multimodal understanding and generation via state space models.
\newblock {\em arXiv preprint arXiv:2503.08686}, 2025.

\end{thebibliography}


\newpage

\appendix

\section{Supplementary Material}

\subsection{More Qualitative Analysis}

\begin{figure*}[htb]
  \centering
  \includegraphics[width=1.0\linewidth]{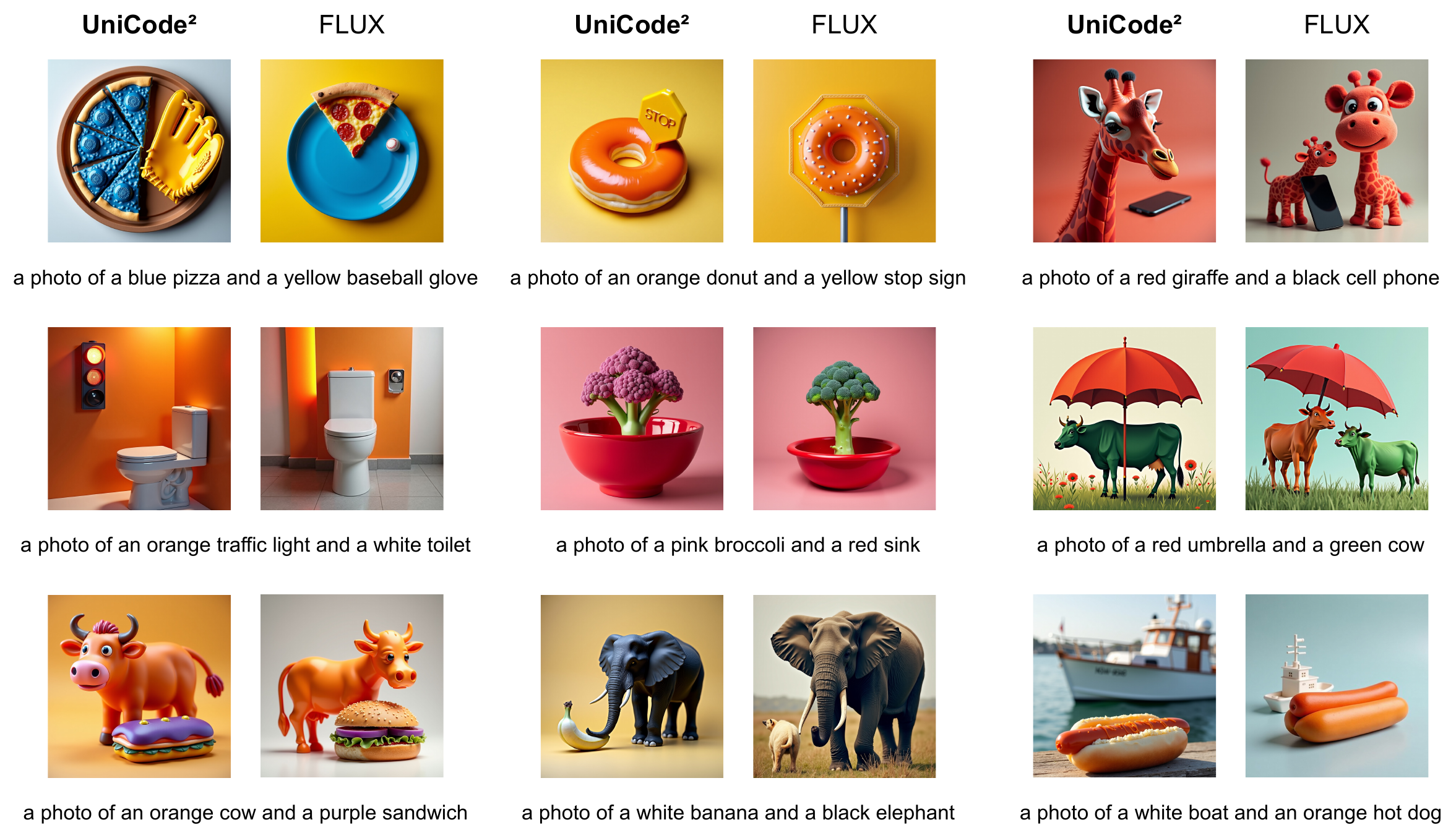}
  \caption{Qualitative comparison between UniCode² and FLUX on text-to-image generation.}
  \label{fig_compare}
\end{figure*}

\paragraph{Qualitative Comparison.} 
Figure~\ref{fig_compare} presents a side-by-side comparison between UniCode² and FLUX~\cite{blackforestlabs2024} on selected prompts from the GenEval benchmark. We observe that UniCode² consistently generates images with better semantic coverage, particularly in cases involving multiple objects, fine-grained attributes, or counter-intuitive compositions (e.g., “a blue pizza”).

This advantage stems from the fact that our visual tokens are not merely structural placeholders but carry semantic grounding learned from SigLIP-aligned codebook construction. As a result, the token sequence encodes rich, compositional semantics that guide the diffusion decoder toward more faithful synthesis. In contrast, baseline methods like FLUX—without semantic token guidance—tend to overfit to frequent visual priors and often ignore rare or contradictory attributes.

While our proposed UniCode² occasionally exhibits imperfect spatial realism (e.g., oversized bananas, or oversized umbrella), it more consistently satisfies the core prompt semantics, demonstrating the effectiveness of our decoding-friendly token design.

\begin{figure*}[htb]
  \centering
  \includegraphics[width=0.7\linewidth]{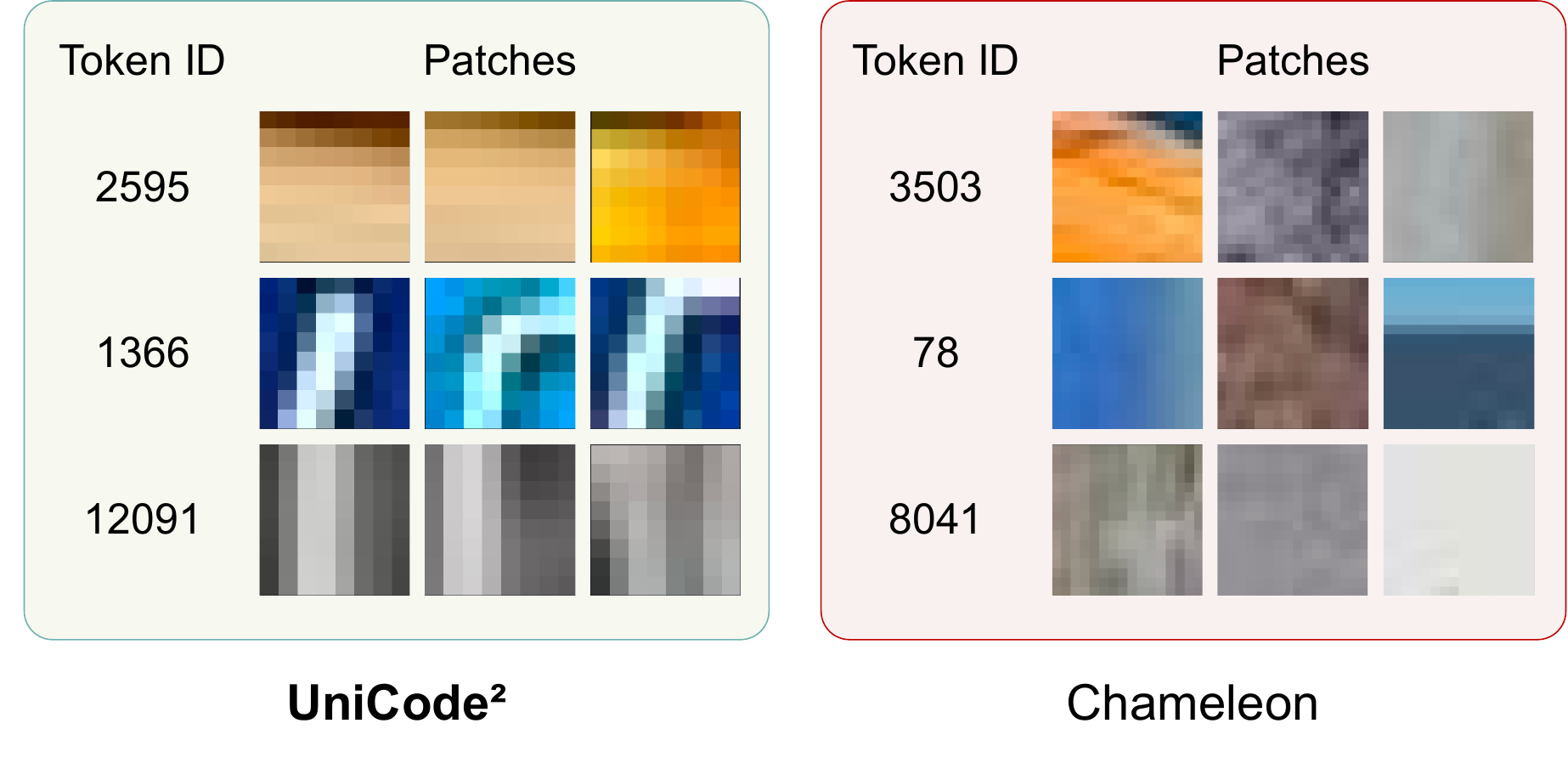}
  \caption{Token-to-patch mapping comparison between UniCode² and Chameleon.}
  \label{fig_vis_tokens}
\end{figure*}

\paragraph{Visualization of Codebook.}
Figure~\ref{fig_vis_tokens} compares token-to-patch mappings between UniCode² and Chameleon\cite{team2024chameleon}. Each row shows image patches assigned to a fixed token ID. We observe that patches grouped under the same token in UniCode² are significantly more semantically consistent—exhibiting shared visual properties such as object identity, color, or texture—whereas Chameleon's tokens tend to cluster less coherent or visually unrelated content.

This improvement arises from our SigLIP-based clustering initialization, which anchors codewords in a semantically aligned feature space, in contrast to Chameleon’s reconstruction-driven tokenization. As a result, UniCode²’s tokens capture higher-level visual abstractions rather than low-level pixel similarities.
Such semantic coherence is beneficial for both understanding and generation: it enables more interpretable and discriminative visual reasoning, and offers stronger conditioning signals during token-to-image synthesis.

\subsection{More Experimental Details}

\paragraph{Large-scale Codebook Construction.}
We construct a 500K-entry visual codebook based on SigLIP-derived semantic embeddings. The process begins by extracting patch-level embeddings from all images used in Stage 1 training (\textasciitilde558K images). For each image, we apply the SigLIP visual encoder and retain all spatial tokens (excluding the [CLS] token), resulting in approximately 150 million patch features in total.

To enable scalable clustering over this massive embedding set, we adopt a two-stage clustering strategy using FAISS. Specifically, we first apply coarse K-means to partition the space into 1K clusters, and then perform fine-grained K-means within each coarse cluster to produce a total of 500K centroids. This hierarchical scheme reduces memory consumption and significantly accelerates convergence compared to directly clustering into 500K classes.

Following clustering, we observe that a small fraction of centroids may contain \texttt{NaN} or \texttt{Inf} values—due to numerical instability when clusters receive very few points. To address this, we remove all invalid centroids and compact the codebook into a contiguous matrix before usage. This filtering step is essential to ensure downstream training stability when embedding lookups and gradient backpropagation are performed over the discrete tokens.

We summarize the construction process in Algorithm~\ref{alg:codebook}.

\begin{algorithm}[htb]
\caption{Large-scale Codebook Construction via Two-Stage Clustering}
\label{alg:codebook}
\begin{algorithmic}[1]
\Require Image set $\mathcal{I}$, visual encoder $\phi$, codebook size $K$, coarse cluster count $K_1$
\Ensure Cleaned codebook $\mathcal{C} \in \mathbb{R}^{K' \times d}$

\State Initialize patch set $\mathcal{P} \gets \emptyset$
\ForAll{image $I \in \mathcal{I}$}
    \State $\mathbf{F} \gets \phi(I)$  \Comment{Extract spatial tokens (excluding CLS)}
    \State Append $\mathbf{F}$ to $\mathcal{P}$
\EndFor

\State \textbf{// Coarse clustering}
\State $\{\mathbf{u}_1, ..., \mathbf{u}_{K_1}\} \gets \text{KMeans}(\mathcal{P}, K_1)$

\State \textbf{// Fine clustering per coarse cluster}
\For{$k = 1$ to $K_1$}
    \State $\mathcal{P}_k \gets$ patches nearest to $\mathbf{u}_k$
    \State $\mathcal{C}_k \gets \text{KMeans}(\mathcal{P}_k, K / K_1)$
\EndFor
\State $\mathcal{C} \gets \bigcup_k \mathcal{C}_k$

\State \textbf{// Filter invalid centroids}
\State $\mathcal{C}_{\text{clean}} \gets \{ \mathbf{c} \in \mathcal{C} : \mathbf{c} \text{ has no NaN/Inf} \}$

\State \Return $\mathcal{C}_{\text{clean}}$
\end{algorithmic}
\end{algorithm}

\paragraph{Training Data.}
To equip UniCode² with unified vision-language understanding and generation, we adopt a multi-stage curriculum learning paradigm inspired by recent MLLM practices~\cite{lu2024ovis, li2024llava, wang2024qwen2, bai2025qwen2}, leveraging the open-source corpus from LLaVA-OneVision~\cite{li2024llava}. 
To train the visual generation pathway, we augment 200K image-caption pairs from JourneyDB~\cite{sun2023journeydb}. These pairs supervise the contrastive alignment (§\ref{sec_3_4}), ensuring semantic consistency of visual tokens.

To equip UniCode² with unified multimodal capabilities across vision-language understanding and generation, we adopt a multi-stage curriculum learning paradigm following recent MLLM practices~\cite{lu2024ovis, li2024llava, wang2024qwen2, bai2025qwen2}. Our training data is based on the open-source corpus released by LLaVA-OneVision~\cite{li2024llava}, consisting of three progressive stages:
\begin{itemize}[leftmargin=*] 
    \item Stage 1: Language-Image Alignment. We use approximately 558K image-caption pairs to align visual representations with the LLM embedding space. This stage focuses on mapping image features into a semantically grounded token space, providing a strong foundation for subsequent multimodal learning.
    \item Stage 1.5: Dense Caption Alignment. To enhance the model's understanding of long-form and compositional semantics, we introduce an additional 4M image-caption pairs featuring richer and denser textual descriptions. These captions contain fine-grained object mentions, spatial relations, and abstract concepts beyond surface-level annotations.
    \item Stage 2: Visual Instruction Tuning. To enable task-specific reasoning and instruction following, we leverage 3.2M high-quality vision-language instruction examples organized into diverse task groups (e.g., VQA, captioning, reasoning, OCR). This stage enhances the LLM's ability to generate aligned and preferred responses across multimodal tasks.
\end{itemize}

For training the visual generation pathway, we supplement Stage 1.5 with an additional 200K image-caption pairs sampled from the JourneyDB~\cite{sun2023journeydb} dataset, which features high-resolution and stylistically diverse images. These examples are used to supervise the contrastive alignment objective (§\ref{sec_3_4}), ensuring that generated visual tokens are semantically consistent and decoding-friendly.

\paragraph{Training Hyper-parameters.}
We summarize the training and inference hyperparameters in Table~\ref{tab:training_hyperparams}.


\begin{table}[htb]
\centering
\small
\setlength{\tabcolsep}{4pt}
\begin{tabular}{lccc}
\toprule
\textbf{Component} & \textbf{Hyper-parameter} & \textbf{Value} & \textbf{Notes} \\
\midrule
\multirow{9}{*}{Training} 
& Learning Rate (Stage 1) & $1 \times 10^{-3}$ & Pre-alignment \\
& Learning Rate (Stage 1.5 / 2) & $1 \times 10^{-5}$ & Instruction tuning \\
& Optimizer & AdamW & $\beta_1{=}0.9, \beta_2{=}0.95$ \\
& Float Precision & bfloat16 & Speed up \\
& Config & zero3 & Deepspeed configuration \\
& Warmup Ratio & 0.03 & Cosine schedule \\
& Batch Size & 512 & Total across GPUs \\
& Epoch & 1 & Total training epoch \\
& Weight decay & 0 & WD setting \\
& GPUs & 128 $\times$ H100 & Device \& amounts \\
\midrule
\multirow{3}{*}{Codebook} 
& Size & 500K & Codebook Size \\
& Clustering Method & FAISS KMeans (2-stage) & $K_1=1\text{K}$ \\
& Input Features & sequence embeddings & siglip-so400m-patch14-384 \\
\bottomrule
\end{tabular}
\caption{Training and model hyperparameters used in UniCode² across different components.}
\label{tab:training_hyperparams}
\end{table}

\subsection{More Explanation on the Motivation}

We expand on the theoretical motivation for jointly optimizing codebook \emph{semantic alignment} and \emph{usage efficiency}. A well-structured visual codebook should minimize quantization error, maximize representational capacity, and maintain stable learning dynamics. We formalize this through four interconnected analyses.

\emph{\textbf{1. Expected Quantization Distortion.}}

Let $\mathcal{E} = \{\mathbf{e}_1, ..., \mathbf{e}_N\} \subset \mathbb{R}^d$ denote visual embeddings sampled from a distribution $p(\mathbf{e})$, and let $\mathcal{C} = \{ \mathbf{c}_1, ..., \mathbf{c}_K \} \subset \mathbb{R}^d$ be a codebook with $K$ discrete centroids. The quantization operator $Q: \mathbb{R}^d \rightarrow \{1, ..., K\}$ maps each embedding to its nearest codeword:
\begin{equation}
Q(\mathbf{e}) = \arg\min_{k \in \{1, ..., K\}} \| \mathbf{e} - \mathbf{c}_k \|^2.
\end{equation}
Let $q(k) := \mathbb{P}(Q(\mathbf{e}) = k)$ denote the marginal assignment distribution, and $p_k(\mathbf{e}) := p(\mathbf{e} \mid Q(\mathbf{e}) = k)$ the conditional density over cluster $k$.

The expected quantization distortion is then:
\begin{equation}
\mathcal{D}_{\text{quant}} := \mathbb{E}_{\mathbf{e} \sim p(\mathbf{e})} \left[ \| \mathbf{e} - \mathbf{c}_{Q(\mathbf{e})} \|^2 \right]
= \sum_{k=1}^K q(k) \cdot \underbrace{\mathbb{E}_{\mathbf{e} \sim p_k} \left[ \| \mathbf{e} - \mathbf{c}_k \|^2 \right]}_{\text{intra-cluster variance}}.
\label{eq:quantization_error}
\end{equation}
This objective favors low intra-cluster variance and dominant assignments. However, without further constraints, the solution may collapse—e.g., all embeddings assigned to a few codewords, leaving the rest underutilized.

\emph{\textbf{2. Entropy Regularization and Trade-off.}}

To prevent such collapse, we regularize the codebook with the Shannon entropy of the assignment distribution:
\begin{equation}
H(q) = -\sum_{k=1}^{K} q(k) \log q(k).
\label{eq:entropy}
\end{equation}
Maximizing $H(q)$ encourages uniform codeword usage, promoting representational diversity and better coverage of the embedding space.

We consider the entropy-regularized quantization loss:
\begin{equation}
\mathcal{L}_{\text{reg}} = \mathcal{D}_{\text{quant}} - \beta H(q),
\label{eq:entropy_reg_loss}
\end{equation}
where $\beta > 0$ controls the trade-off. This formulation is closely related to deterministic annealing, and leads to optimal clustering behavior when $\beta$ is annealed appropriately. The minimizer of this loss balances semantic fidelity and codeword diversity.

\emph{\textbf{3. Information-Theoretic Capacity.}}

Let $y$ be a semantic target label (e.g., object class, scene type), and let $v = Q(\mathbf{e})$ be the discrete token assigned to embedding $\mathbf{e}$. The mutual information between tokens and labels satisfies:
\begin{equation}
I(y; v) = H(v) - H(v \mid y) \leq H(v) = H(q),
\label{eq:mutual_info_bound}
\end{equation}
where equality holds only if token assignments are fully deterministic given labels. Thus, a codebook with low entropy fundamentally limits the maximal mutual information, and thereby the ability to distinguish between semantic classes.

Moreover, from the information bottleneck perspective, visual tokens should preserve as much information about $y$ as possible while compressing $\mathbf{e}$. The constrained bottleneck objective:
\begin{equation}
\min_{Q} \; \mathbb{E}[\| \mathbf{e} - \mathbf{c}_{Q(\mathbf{e})} \|^2] \quad \text{s.t.} \quad I(y; v) \geq \tau
\end{equation}
is only satisfiable if $H(q)$ is large enough to support high-fidelity compression of semantic variation.

\emph{\textbf{4. Asymptotic Distortion Bound and Uniformity.}}

In classical high-resolution quantization theory, the optimal rate-distortion function satisfies:
\begin{equation}
\mathcal{D}_{\text{quant}} \geq C_d \cdot K^{-2/d},
\end{equation}
where $C_d$ depends on the intrinsic dimension of the embedding space. Crucially, this bound is only achievable under two conditions:
(1) the data lies on a smooth manifold, and (2) codewords are uniformly spaced and equally used.

This confirms that both semantic compactness (low intra-cluster variance) and codeword balance (high $H(q)$) are \emph{necessary} for asymptotic optimality. In practice, standard randomly initialized codebooks (e.g., VQ-GAN) fail to satisfy either, especially at large $K$.

\emph{\textbf{5. Summary and Design Implication.}}

Together, these results show that an optimal codebook should simultaneously minimize semantic distortion and maintain high usage entropy:
\begin{equation}
\mathcal{C}^\ast = \arg\min_{\mathcal{C}} \left\{ \sum_{k=1}^K q(k) \cdot \mathbb{E}_{p_k}[\| \mathbf{e} - \mathbf{c}_k \|^2] - \beta H(q) \right\}.
\label{eq:final_objective}
\end{equation}
Our SigLIP-based codebook construction approximates this objective by clustering in a semantically aligned space with high initial entropy. As a result, we are able to scale to 500K tokens with near-uniform utilization and strong performance across understanding and generation tasks.






\end{document}